%% file: main.tex
\documentclass[letterpaper]{article} % DO NOT CHANGE THIS
\usepackage{aaai23}  % DO NOT CHANGE THIS
\usepackage{times}  % DO NOT CHANGE THIS
\usepackage{helvet}  % DO NOT CHANGE THIS
\usepackage{courier}  % DO NOT CHANGE THIS
\usepackage[hyphens]{url}  % DO NOT CHANGE THIS
\usepackage{graphicx} % DO NOT CHANGE THIS
\usepackage{natbib}  % DO NOT CHANGE THIS AND DO NOT ADD ANY OPTIONS TO IT
\usepackage{caption} % DO NOT CHANGE THIS AND DO NOT ADD ANY OPTIONS TO IT
\usepackage{algorithm}
\usepackage{algorithmic}
\usepackage{lipsum}
\usepackage[utf8]{inputenc} % allow utf-8 input
\usepackage[T1]{fontenc}    % use 8-bit T1 fonts
\usepackage{url}            % simple URL typesetting
\usepackage{booktabs}       % professional-quality tables
\usepackage{amsfonts}       % blackboard math symbols
\usepackage{nicefrac}       % compact symbols for 1/2, etc.
\usepackage{microtype}      % microtypography
\usepackage{xcolor}         % colors
\usepackage{array,multirow,graphicx}
\usepackage{float}
\usepackage{amsmath}
\usepackage{mathtools}
\usepackage{dsfont}
\usepackage{hhline}%#######
\usepackage{boldline}

\DeclarePairedDelimiter\floor{\lfloor}{\rfloor}
\usepackage{pifont}% http://ctan.org/pkg/pifont
\usepackage{soul} %\st
\usepackage{algorithm}
\usepackage{verbatim}
\usepackage{paralist}
\usepackage{algorithmic}
\usepackage{newfloat}
\usepackage{listings}

\newcommand{\cmark}{\ding{51}}%
\newcommand{\xmark}{\ding{55}}%

\usepackage{newfloat}
\usepackage{listings}
\DeclareCaptionStyle{ruled}{labelfont=normalfont,labelsep=colon,strut=off} % DO NOT CHANGE THIS
\floatstyle{ruled}
\newfloat{listing}{tb}{lst}{}
\floatname{listing}{Listing}
\title{Minority-Oriented Vicinity Expansion with Attentive Aggregation\\ for Video Long-Tailed Recognition}
\author{
    WonJun Moon,
    Hyun Seok Seong,
    Jae-Pil Heo\thanks{Corresponding author}
}
\affiliations{
    Sungkyunkwan University\\
    \{wjun0830, gustjrdl95, jaepilheo\}@skku.edu
}

\newcommand{\Skip}[1]
{
}
\usepackage{bibentry}
\begin{document}

% \bibliography{aaai23}
\maketitle
\input{0_abstract}
\input{1_introduction}

\input{2_aggregation}
\input{3_method}

\input{4_experiment}
\input{5_related_work}
\input{6_conclusion}
{\small
    % \bibliographystyle{plainnat}
    % plainnat abbrvnat unsrtnat
    \bibliography{aaai23}
}
\newpage
\input{appendix_0_overallflow}

\input{appendix_1_implementation}
\input{appendix_2_ablation}

\input{appendix_3_data}
\end{document}

%% file: 0_abstract.tex
%Balancing the distribution between categories is challenging as the real-world video volume is being increased dramatically.
%This spotlights the need for Video Long-Tailed Recognition~(VLTR) to learn from long-tailed distribution.
\begin{abstract}
A dramatic increase in real-world video volume with extremely diverse and emerging topics naturally forms a long-tailed video distribution in terms of their categories, and it spotlights the need for Video Long-Tailed Recognition (VLTR).
In this work, we summarize the challenges in VLTR and explore how to overcome them.
The challenges are: (1) it is impractical to re-train the whole model for high-quality features, (2) acquiring frame-wise labels requires extensive cost, and (3) long-tailed data triggers biased training.
Yet, most existing works for VLTR unavoidably utilize image-level features extracted from pretrained models which are task-irrelevant, and learn by video-level labels.
Therefore, to deal with such (1) task-irrelevant features and (2) video-level labels, we introduce two complementary learnable feature aggregators.
Learnable layers in each aggregator are to produce task-relevant representations, and each aggregator is to assemble the snippet-wise knowledge into a video representative.
Then, we propose Minority-Oriented Vicinity Expansion (MOVE) that explicitly leverages the class frequency into approximating the vicinity distributions to alleviate (3) biased training.
By combining these solutions, our approach achieves state-of-the-art results on large-scale VideoLT and synthetically induced Imbalanced-MiniKinetics200. 
With VideoLT features from ResNet-50, it attains 18\% and 58\% relative improvements on head and tail classes over the previous state-of-the-art method, respectively. 
% Our code is available at https://github.com/wjun0830/MOVE.
\end{abstract}

%% file: 1_introduction.tex
\section{Introduction}
% 0. Video recognition
With the development of online video platforms, recent years have witnessed a rapid growth in interest in learning from video.
However, unlike the image domain, video has an additional temporal dimension which requires huge memory and computation cost.
Thus, it has been an inevitably common approach to utilize the extracted features using pretrained networks.~\cite{brattoli2020rethinking, xu2020g, lin2019bmn}.
% However, this raises the issue of using task-irrelevant features due to the domain gap between datasets to pretrain the network and be evaluated in downstream tasks.
However, this raises the issue of using task-irrelevant features in downstream tasks due to the domain gap between datasets to pretrain the network and be utilized.% for target tasks.
VLTR is not an exception since its focus is to learn from a vast amount of videos being collected via online platforms.
% Therefore, although the domain gap between datasets leads to the use of task-irrelevant features for downstream tasks, it has been an inevitably common approach to utilize the extracted features using pretrained networks~(\cite{feng2021mist, videolt, brattoli2020rethinking, xu2020g, lin2019bmn}).
% Therefore, it has been an inevitably common approach to utilize the extracted features using pretrained networks despite the domain gap between datasets~(\cite{feng2021mist, videolt, brattoli2020rethinking, xu2020g, lin2019bmn}).
% 1. popular/practical long tail도 이런 문제가 있음.
% 게다가, image long tail을 바로 적용하느 ㄴ것도 어려움.
% 그래서 long tail해결하려면 이것도 저것도 동시에 해결해야함.

\begin{figure*}
    \centering
    \includegraphics[width =0.82\textwidth]{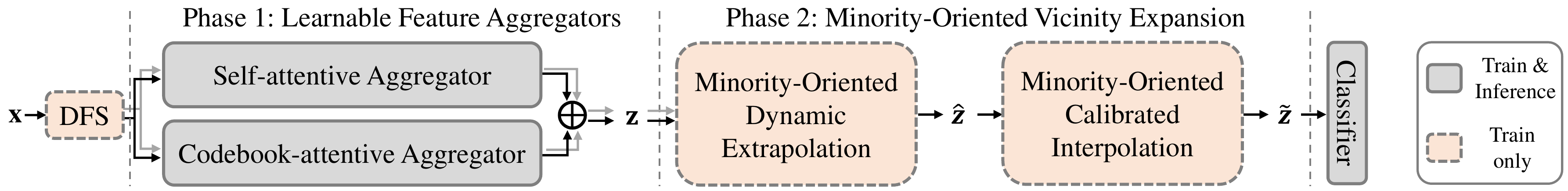}
    % \vspace{-0.1cm}
    \caption{
    Overall flow. Given input feature $\mathbf{x}$, our proposed Dynamic Frame Sampler (DFS) is firstly performed as an effective augmentation tool when developing a video prototype~(Sec.\ref{Sec.3.2.TOMix}). These features are condensed into task-relevant video representation by two aggregators~(Sec.\ref{Sec.2.LearnableFeatureAggregator}). After concatenation, a balanced training of the classifier based on expanding and extending the representation space of the minorities is led by our proposed MOVE~(Sec.\ref{Sec.3.2.TOMix}). 
    At inference, only the learnable feature aggregators and the classifier are used for prediction.
    }
    \label{Fig.trainingflow}
    % \vspace{-0.2cm}
\end{figure*}

Data imbalance, a natural phenomenon in these real-world datasets, is detrimental since it triggers biased training that results in the model performing poorly on tail classes~\cite{GCL, tadelongtail}.
To overcome the difficulty, there were numerous approaches in the image domain such as reweighting and resampling~\cite{cao2019learning, fan2017learning, menon2020long, ren2020balanced, shu2019meta}.
However, directly applying these techniques to the video domain is impractical because acquiring accurate frame-wise supervision for massive video datasets is usually cumbersome, and some snippets are no more than background frames~\cite{videolt}.
Thus, data imbalance in the video domain should be considered with the issue of learning from task-irrelevant features and weakly-labeled problems that only the video-level label is available.
To this end, we first employ learnable feature aggregators to rectify and aggregate the frame-wise representations at a video-level where the weak labels are available.
Previously, Framestack~\cite{videolt} stuck with the frame-level balancing technique since not all snippets contain informative clues.
Nevertheless, our intuition is that snippet-wise feature in weakly labeled settings is detrimental since the precise labels cannot be assigned.
% Nevertheless, our intuition is that snippet-wise label utilization in weakly labeled settings usually requires model's generalization capability to proofread the weak labels~(\cite{feng2021mist, zhong2019graph}).
Label cleaning~\cite{feng2021mist, zhong2019graph} that generates reliable snippet-wise pseudo labels, is also impractical since neural networks are poor at generalizing on the minority.
Therefore, we argue that weakly labeled VLTR should be treated at the video-level where noisy background scenes are watered down. % , thereby negligible.
% However, deep neural networks~(DNN) are poor at generalizing on the minority.
% Nevertheless, our intuition is that filtering frame-level noise labels in tail classes is difficult because deep neural networks~(DNN) are poor at generalizing on tail classes.
% Therefore, we argue that weakly labeled VLTR should be treated at the video level where noisy background scenes are watered down thereby negligible.
Subsequently, we propose Minority-Oriented Vicinity Expansion~(MOVE) to prevent the majority-biased training and overfitting to the partial data of the tail classes.
Specifically, we expand the empirical distribution of sparse minorities by dynamic extrapolation and extend the distribution by assigning more linearly designed space between the class distributions to the minority than the majority with calibrated interpolation.
Overall, our framework comprises two phases as shown in Fig.~\ref{Fig.trainingflow}. 
% We visit each phase in Sec.~\ref{Sec.2.LearnableFeatureAggregator} and Sec.~\ref{Sec3.MOVE}.
% After discussing each phase, we revisit the overall flow.
Our key contributions are:
\begin{itemize}
    \item We summarize the challenges in VLTR and empirically verify the importance of handling these challenges with our ablation studies. We believe that our findings clearly segment and introduce the research direction.    
    \item We introduce learnable feature aggregators as an effective tool to obtain task-relevant and label-consistent features in weakly-labeled VLTR. Experiments demonstrate that baselines are significantly improved with our aggregators.
    % \item We introduce learnable feature aggregators as an effective tool to obtain task-relevant and label-consistent features in weakly-labeled VLTR. 
    % Experiments demonstrate that baselines are significantly improved with our aggregators.
    % Our experiments also demonstrate that existing baselines can also be significantly improved with learnable aggregators.
    \item We propose Minority-Oriented Vicinity Expansion to generalize the model to long-tailed data distribution by alleviating the biased training. 
    % Experiments validate the effectiveness of alleviating the biased training with long-tailed data.
    \item Our overall approach is verified to be superior to existing state-of-the-art methods with extensive experiments.
    % \item \rebuttal{We introduce artificially synthesized dataset, Imbalanced-MiniKinetics, which provides diverse benchmarking scenarios that are controlled by different imbalance ratios.}
    % \item We newly introduce an imbalanced video classification benchmark, Imbalanced-MiniKinetics200, sampled from Kinetics-400 to validate the performance of our method in diverse imbalanced classification scenarios.
    \item We newly introduce an imbalanced video classification benchmark, Imbalanced-MiniKinetics200, sampled from Kinetics-400 to evaluate diverse imbalanced scenarios.
\end{itemize}

%% file: 2_aggregation.tex
\section{Learnable Feature Aggregator}
\label{Sec.2.LearnableFeatureAggregator}

\subsection{Background and problems}
Oftentimes, it is unrealistic to re-train the entire network to get access to the quality of feature maps.
It is especially true in the video domain because of the huge size of the videos.
Thus, it has been common to use popular pretrained backbones.
Likewise, VLTR also aims to learn from real-world data collected day and night.
% This gains more credibility in the video domain since its objective is to learn from massive real world data that is collected day and night.
% Thus, popular pretrained backbones are often employed to reduce the data dimension.
% Particularly, this gains more credibility importance since long-tailed recognition focuses on learning from massive real world data that is collected day and night.
Yet, video recognition obviously depends on the quality of class-discriminative features while pretrained networks may not fit for extracting such quality due to the domain gap between datasets~\cite{choi2020unsupervised}.

In addition, processing snippet-wise features require accurate snippet-wise label information that is accompanied by extensive human labor.
To extract reliable supervision in a weakly labeled setting, label cleaning has been proposed where the model iteratively assigns and corrects the snippet-wise labels with its predictions~\cite{feng2021mist, zhong2019graph}.
However, under the circumstance where the model's predictions are not precise, especially on tail classes, it is difficult to proofread the label precisely.
Therefore, we argue that it is better to modify snippet-wise features to match the degree of trustworthy supervision at the video-level rather than the other way around, e.g., label cleaning.
In short, feature aggregation addresses the level mismatch between given snippet-wise feature representation and video-wise supervision, alleviating the label uncertainty.

\begin{figure*}[t]
    \centering
    \includegraphics[width=0.80\textwidth]{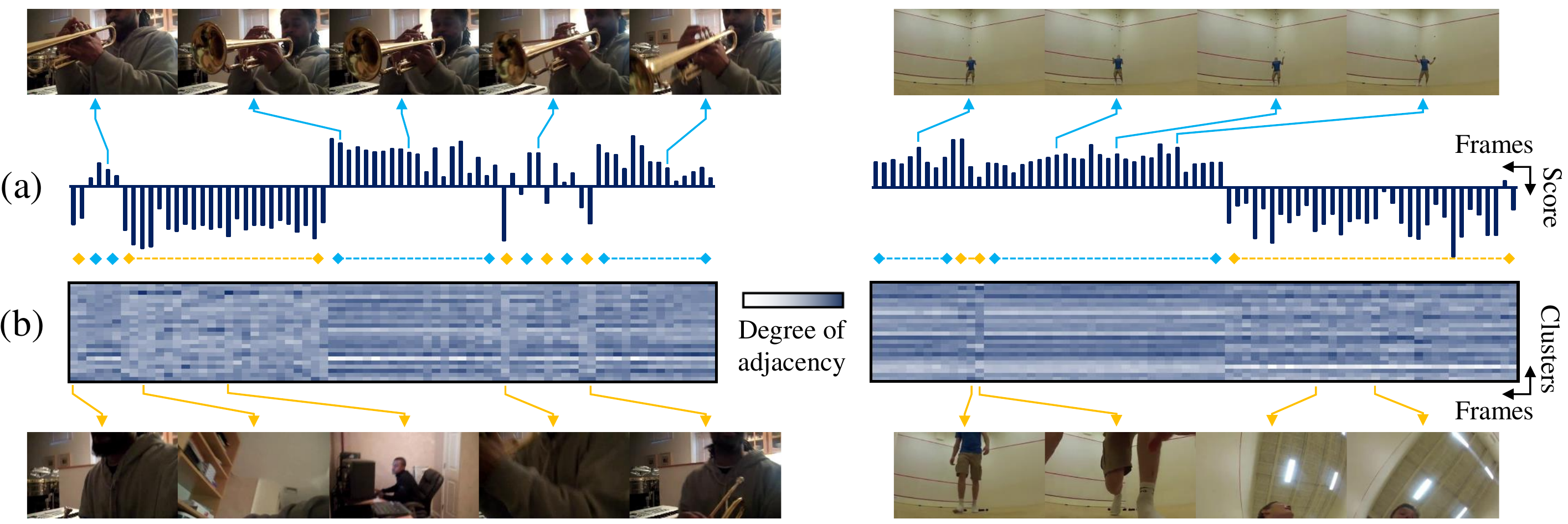}
    % \vspace{-0.2cm}
    \caption{Capability of localizing class-discriminative frames of two aggregators introduced in Sec.~\ref{Sec.2.LearnableFeatureAggregator}. 'trumpet performance' and 'jugglingBall' classes in VideoLT are used, and the X-axis of (a) and (b) are shared.
    \textbf{(a)} Per-frame attention score~($\textit{QK}^\mathsf{T}/\sqrt{d}$) in self-attentive aggregator.
    As high attention scores are given to frames deeply involving class information~(blue arrow) and low scores on noisy frames~(yellow arrow), it is obvious that the video prototype feature is well supplemented with class-relative local clues.
    % In Sec.~\ref{Sec.4.Ablation}, we further compare how our self-attentive aggregator achieves more gains than the naive use of MSA.
    In Sec.~\ref{Sec.4.Ablation}, we further analyze how our self-attentive aggregator achieves more gains than the naive use of MSA by addressing two subproblems in VLTR.
    % This leads to successful mitigation of label uncertainty.
    \textbf{(b)} Per-frame degree of adjacency to each cluster in codebook-attentive aggregator.
    % S-A gives a high attention score to frames with deeply involving class information~(blue arrow), and a thoroughly low score to noisy frames~(yellow arrow).
    It can be noticed that degree of adjacency to all clusters reflects the frame similarity. Similar frames are assigned with similar adjacency values~(blue dotted lines), whereas others are not~(yellow dotted lines).
    % It can be noticed that similar frames are assigned with similar adjacency~(blue dotted lines) maps, whereas noisy frames have different maps~(yellow dotted lines).
    % C-A assigns almost identical adjacency group for class-discriminative frames~(frames in blue lines), and noisy adjacency group for noisy frames~(frames in yellow lines).
    % Note that the class-discriminative frames activated in two aggregators, (a) and (b), are identical to each other~(frames in blue lines).
    % This leads to successful mitigation of label uncertainty.
    These scores clearly describe how well our aggregators understand and represent the video.
    }
    % \vspace{-0.4cm}
    \label{fig_attention_score}
\end{figure*}

\subsection{Generating task-relevant and label-consistent representation}
% Prior to handling data with long-tailed distribution, 
With features from pretrained networks, we aim to produce task-relevant discriminative representations and address a weakly labeled problem when it concurrently exists with data imbalance.
To manage these, we introduce two learnable feature aggregators: self-attentive and codebook-attentive aggregators.
Specifically, the role of learnable layers within these aggregators is to develop task-relevant representations overcoming the domain gap between the dataset to pretrain the network and the target dataset for the downstream task.
Label uncertainty from weak labels is also relieved by the aggregation process to equalize the level of data point and its supervision.
Note that the loss of information while feature compression, is not critical since two complementary aggregators alleviate the issue and a well-chosen prototype is often enough to achieve strong performance~\cite{buch2022revisiting}.

% These aggregators implement this in a complementary way when compressing temporal dimensions, which can mitigate the loss of information triggered by aggregation.
% To manage these, we introduce two reciprocal feature aggregators to yield task-relevant and label-consistent representation: self-attentive and codebook-attentive.

\textbf{Self-attentive aggregator.}
Self-attention~(SA)~\cite{SA, VIT} aggregates feature maps with normalized importance that indicates the relationship between a token and all other tokens. Specifically, given $i$-th input feature $\mathbf{x}^i$ from a pretrained network to linearly project each query $\textit{Q}$, key $\textit{K}$, and value $\textit{V}$, SA operates as follows:
\begin{equation}
\label{eq_SA}
    % \mathbf{\hat{x}}^i = \text{Softmax}\big(\frac{\textit{QK}^\mathsf{T}}{\sqrt{d}}\big)\textit{V},
    \mathbf{\hat{x}}^i = \text{Softmax}\big(\textit{QK}^\mathsf{T} / {\sqrt{d}}\big)\textit{V},
\end{equation}
where $d$ is the feature dimension of feature $\mathbf{x}^i$.
Multihead Self-Attention (MSA) is an extension of SA where multiple heads exist to implement self-attention operations.
Such MSA can be interpreted as an ensemble over SA for searching robust attention maps by considering different representations.

With the property of local attention, our focus is to efficiently generate video-level representation. 
Therefore, we simply modify MSA to introduce Prototypical Self-Attention (PSA) that complements the global feature with the snippet-level local clues.
In PSA, we project Pool$(\mathbf{x}^i)$ into $\textit{Q}$ instead of projecting $\mathbf{x}^i$ where Pool$()$ denotes the function for average pooling to aggregate temporal information.
Then, global feature, Pool$(\mathbf{x}^i)$, is supplemented with the informative local clues from frame-wise features $\mathbf{x}^i$ to represent the video instance.
% To supplement global feature with the informative clues in local frames and make a video prototype, we project $\mathbf{x}_{global}^i$ to $\textit{Q}$ and $\mathbf{x}_{local}^i$ to $\textit{K, V}$ in Eq.~\ref{eq_SA}.
As shown in Fig.~\ref{fig_attention_score}, meaningful local clues are highly likely to be assembled to the prototype, while frames with class-meaningless features are not. %filtered out
Consequently, PSA yields $\mathbf{\hat{x}}^i$, the class-discriminative prototype at the same level as weak labels.
Additionally, it has its benefits in efficiency since only the prototype is re-expressed with other snippets.
% For the efficiency, PSA only requires FLOPS about half the MSA since only the prototype is expressed with other snippets.%(48\%)
% generates a  feature by considering temporally local relationships which balances the degree of 
% at the same level as the weak labels.
% By integrating instructive features into prototypical representation, video-descriptive representations are at the same level as the weak labels.
% Then, $z_{global}$ is processed to calculate the $\textit{Q}$ in Eq.~\ref{eq_SA}. 
% \begin{equation}
% \label{eq_SA}
%     \hat{z} = \textit{Softmax}\big(\frac{z_{global} \cdot z}{\sqrt{d}}\big)z,
% \end{equation}
% As a result, global feature is supplemented with the informative clues in local frames in which becomes global-local features which clearly represent the WS labels at a video-level.
% Manipulating global-local features also has its benefits on computation complexity by requiring only 48\% floating operations compared to directly applying MSA on the temporal axis.

\textbf{Codebook-attentive aggregator.}
NetVLAD~\cite{netvlad}, the differentiable form of VLAD~\cite{vlad}, is another strategy to aggregate a set of features effectively.
Generally, with the feature representation $\mathbf{x}^i \in \mathbb{R}^{T \times C}$ from a video stream as an input, VLAD assumes $K$ clusters $\boldsymbol{\mu} \in \mathbb{R}^{K \times C}$ in the global codebook.
With the residues from these two vectors, VLAD outputs $C \times K$-dimensional representation $\tilde{\mathbf{x}}^i$:
\begin{equation}
\label{eq_vlad}
    \mathbf{\tilde{x}}^i_{c, k} = \sum^{T}_{t=1}\rho_{k}(\mathbf{x}^i_{t})(\mathbf{x}^i_{t,c}-\boldsymbol{\mu}_{k, c}),
\end{equation}
where $\rho_{k}(\mathbf{x}_{t}^i)$ is an indicator function that checks if the $k$-th cluster, $\boldsymbol{\mu}_k$, is the closest cluster from $\mathbf{x}_t^i$, the $t$-th component in the temporal dimension. $\mathbf{x}_{t,c}^i$ and $\boldsymbol{\mu}_{k,c}$ denote $c$-th dimension of $\mathbf{x}_t^i$ and $\boldsymbol{\mu}_k$, respectively.
To make it differentiable, NetVLAD proposed to use softmax function for $\rho_k(\mathbf{x}_t^i)$ as:
\begin{equation}
\label{eq_netvlad}
    \rho_k(\mathbf{x}^i_t) = \frac{e^{w_k^\mathsf{T}\mathbf{x}_t^i+b_k}}{\sum^{K}_{{k}^\prime=1}e^{w_{{k}^\prime}^\mathsf{T}\mathbf{x}^i_t+b_{k^\prime}}},
\end{equation}
where $w_k$, $b_k$, and $\boldsymbol{\mu}_k$ are trainable parameters for $k$-th cluster in NetVLAD, respectively.
Since the operation has the property of sharing the global codebook to express every video, we employ it as a codebook-attentive aggregator to represent the global relationship.
Note that, although NetVLAD itself is not a part of our contribution, we believe that utilizing it together with the self-attentive aggregator to minimize information loss by complementing each other and the following discussion has a worthwhile technical contribution.
%Remark that the usage of NetVLAD is not our contribution; however, our contribution here lies in the objective of supplementing the self-attentive aggregator to minimize the information loss while aggregating and analyzing how it supplements in a complementary way in the discussion.
In Fig.~\ref{fig_attention_score}, the video understanding capability of the codebook-attentive aggregator is demonstrated as similar scenes exhibit a similar degree of adjacency to each cluster.

% In order to show that the codebook-attentive aggregator is ab

% In order to show the  effectiveness in localizing class-descriptive representation, we show how the codebook can cluster class-descriptive frames with its score.
% As the purpose is same as self-attentive aggregator, we also show how class-relative frames are represented similarly and others different with the scores of the codebook.

\textbf{Discussion.}
Whereas the self-attentive aggregator considers internal relationships within videos to make class-relative representation, the codebook-attentive aggregator utilizes global relationships with clusters to implement the fine-grained discrimination beyond the class level.
In Fig.~\ref{fig_attention_score}, it is shown that attention scores in (a) are computed solely within a video instance (i.e., the relevance of each frame with respect to a video) to highlight task-relevant frames. 
On the other hand, the scores in (b) are defined with a codebook shared by all the instances to produce more fine-grained classification-friendly representations.
% Thus, even when the noisy prototype is yielded from self-attentive aggregator due to dominant backgrounds, it is not critical since it is relieved by codebook-attentive prototype.
In other words, it is rather straightforward that each module pursues a different way of deriving task-relevant and label-consistent representation, thereby supplementing each other (e.g., the codebook-attentive aggregator can complement the self-attentive aggregator when dominant backgrounds form noisy prototype in PSA).
% and the other way around when similar identities exist between classes.
% In other words, it is rather straightforward that each module pursues a different way of deriving task-relevant and label-consistent representation.
Thus, we define the final aggregated feature as:
% \textbf{Discussion.}
% Whereas the self-attentive aggregator considers internal relationships within videos to make class-level discriminative representation, the codebook-attentive aggregator utilizes global relationships with clusters to implement the fine-grained discrimination beyond the class-level.
% In other words, it is rather straightforward that each module pursues a different way of deriving task-relevant and label-consistent representation.
% Thus, we define the final aggregated feature as:
\begin{equation}
\label{eq_finalAgg}
    \mathbf{z}^i = f_\theta(\mathbf{\hat{x}}^i) \oplus f_\phi(\mathbf{\tilde{x}}^i),
%     \dot{f}(\dot{z})
% \ddot{f}(\ddot{z})
\end{equation}
where $f_\theta(\cdot)$ and $f_\phi(\cdot)$ are fully connected layers parameterized by $\theta$ and $\phi$, respectively, to project into a reduced dimension. $\oplus$ stands for concatenation.
By combining these features, we achieve task-relevant and label-consistent video representation without much loss of information.
% Whereas the self-attentive aggregator consider internal relationships within videos to make class-level discriminative representation, codebook-attentive aggregator focuses on the instance-level fine-grained discrimination.
% concatenating two latent representations can be understood as generating discriminative 
% It can also be understood as representations from one another can be enriched by considering the other property. 
% Hence, we concatenate two latent features from each aggregator to form a new latent space that considers both internal and external relationships.

% Without the need to discourage the correlation between two aggregators, it is rather straightforward that each module has different properties since they only consider either internal relationships within videos or external relationships between videos.
% since they only consider either internal relationships within videos or external relationships between videos.
% whereas the main objective of self-attentive aggregator is to decrease the label uncertainty, codebook-attentive aggregator instance-level fine-grained discrimination
% It can also be understood as representations from one another can be enriched by considering the other property. 
% Hence, we concatenate two latent features from each aggregator to form a new latent space that considers both internal and external relationships.
% Local / Global attention

%% file: 3_method.tex
\section{Minority-Oriented Vicinity Learning}
\label{Sec3.MOVE}

As described in Sec.\ref{Sec.2.LearnableFeatureAggregator}, we propose and adopt aggregators to produce task-relevant representations and reduce the label uncertainty by uniting frame-level features into the prototype.
Meanwhile, these do not provide a solution for balancing the biased training phrase with the long-tailed data distribution.
Since it is widely known that the long-tailed distribution triggers the learning of head-biased decision boundaries~\cite{GCL, ibloss, kang2019decoupling, mislas}, we focus on readjusting such distorted boundaries.
Still, there are insufficient samples for specific tail classes to achieve the adjustment.
Therefore, we instead minimize the estimated vicinal risk from the calibrated vicinity to approximate the real data distribution.
% Since it is widely known that the long-tailed distribution triggers to form head-biased boundaries~(\cite{GCL, ibloss, kang2019decoupling, mislas}), we focus on readjusting distorted boundaries.
% Still, there are insufficient samples for specific classes.
% Therefore, we address the problem by minimizing the calibrated vicinal risk to approximate the real data distribution.
% Still, as the quantity of samples are not available for every class, we instead minimize the vicinal risk to approximate the real data distribution and calibrate the boundaries with them.
% To unfold the head-biased decision boundaries by diversifying given minorities, we propose Minority-Oriented Density Estimate Learning (MODEL) which manipulates both the representation space and the label space to form a minority-rich density estimate.

\textbf{Vicinal risk minimization.} Given $Z$ and $Y$, a set of $i$-th input vector $\mathbf{z}^i$ and corresponding one-hot label $\mathbf{y}^i$, the objective of supervised learning is to find a hypothesis $h$ that $h : Z \to Y$.
% Note that $\mathbf{z}$ in this section indicates the aggregated feature $\mathbf{x}^{AGG}$.
However, due to the unavailability of true distribution, the expected risk for learning $h$ cannot be computed.
Although it is a natural practice to use empirical risk minimization instead, it is highly dependable on the quality of given data distribution.
Then, \cite{chapelle2000vicinal} proposed Vicinal Risk Minimization~(VRM) that approximates the expected risk with below equation:
\begin{align}
\label{eq_vrm}
    R_{vic}(h) &= \frac{1}{N}\sum_{i=1}^{N}\int\ell(h(\mathbf{z}), \mathbf{y}^{i})dP_{\mathbf{z}^{i}, \mathbf{y}^{i}}(\mathbf{z}, \mathbf{y}) \\ & = \frac{1}{N}\sum_{i=1}^{N}\ell(h(\tilde{\mathbf{z}}^{i}), \tilde{\mathbf{y}}^{i}),
\end{align}
% \begin{equation}
% \label{eq_vrm}
%     R_{vic}(h) = \frac{1}{N}\sum_{i=1}^{N}\int\ell(h(\mathbf{z}), \mathbf{y}^{i})dP_{\mathbf{z}^{i}, \mathbf{y}^{i}}(\mathbf{z}, \mathbf{y}),
% \end{equation}
where $P_{\mathbf{z}^{i}, \mathbf{y}^{i}}(\mathbf{z}, \mathbf{y})$ indicates a distribution in the vicinity of the data point ($\mathbf{z}^{i}, \mathbf{y}^{i}$), and ($\tilde{\mathbf{z}}^i, \tilde{\mathbf{y}}^{i}$) is a data point sampled from $P_{\mathbf{z}^{i}, \mathbf{y}^i}(\mathbf{z}, \mathbf{y})$.
$N$ is the total number of training data.
Simply put, the goal of VRM is to address the problem of missing data by discovering new kinds of patterns with the quality of density estimates.
Therefore, we tailor VRM to the long-tailed problem by modeling the vicinity within/between samples and manipulating with the consideration of long-tailed sample gaps.
We show how the vicinal distribution $P_{\mathbf{z}^{i}, \mathbf{y}^{i}}(\mathbf{z}, \mathbf{y})$ is defined by MOVE in the following subsection.

\begin{figure}
    \centering
    % \vspace{-0.2cm}
    % \includegraphics{fig/figure_embedding_illustration_v4.pdf}
    \includegraphics[width =0.47\textwidth]{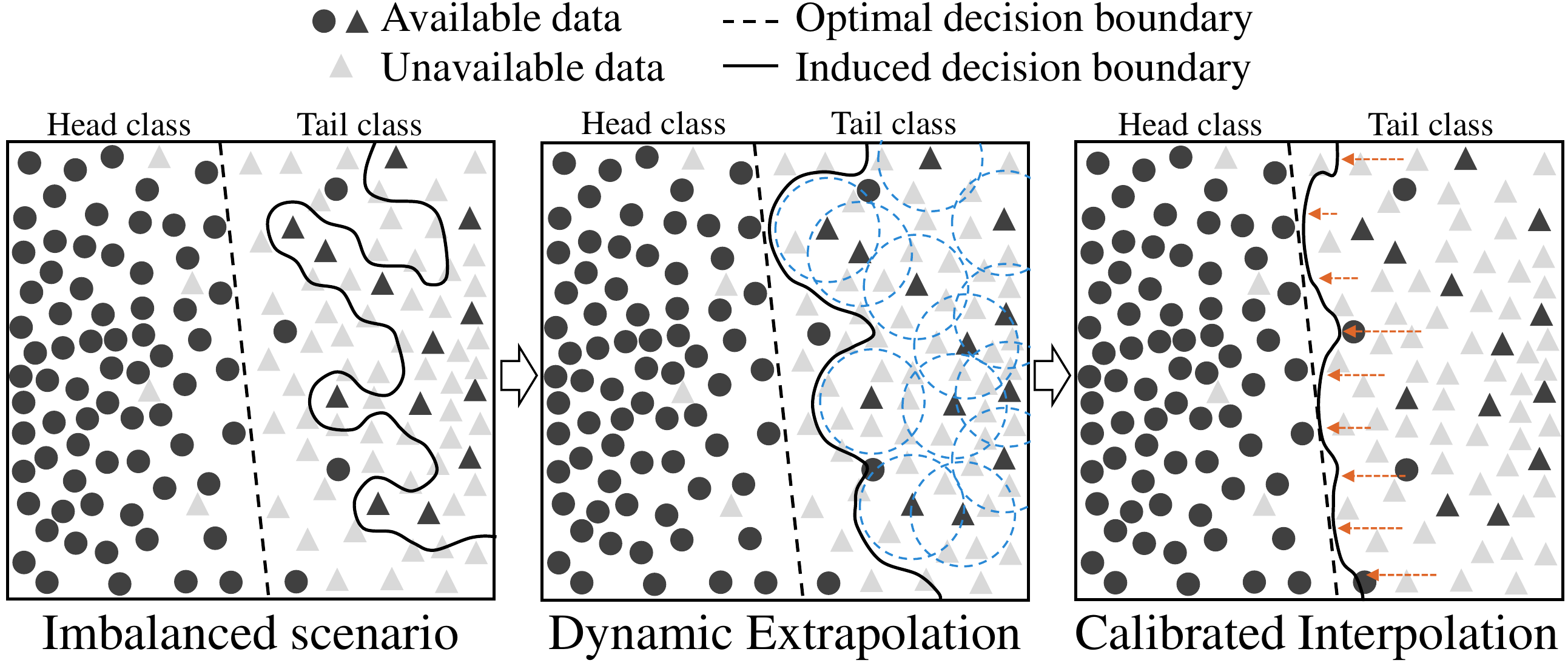}
    % \vspace{-0.5cm}
    \caption{
    The key concept of MOVE.
    \textbf{(Left)} In imbalanced scenarios, boundaries are shifted and distorted towards the minority.
    \textbf{(Center,~Right)} We propose two components of MOVE, namely dynamic extrapolation, and calibrated interpolation, that aim to smooth and balance the boundary by expanding and extending the tail vicinity distribution.
    DFS guides the extrapolation and interpolation to be focused on the minority.
    % Illustration of changes in decision boundary for Minority-Oriented Density Estimate Learning.
    }
    % \vspace{-0.4cm}
    \label{fig_embedding_illustration}
\end{figure}
% We show how the vicinal data point~($\tilde{\mathbf{z}}_i$, $\tilde{\mathbf{y}}_i$) is defined for Minority-Oriented Vicinity Expansion in the following subsection.
% Therefore, we employ VRM to seek and optimize diversified minority samples with the consideration of long-tailed sample gaps in both the embedding and the label space.

\subsection{Minority-Oriented Vicinity Expansion}
\label{Sec.3.2.TOMix}
Under the axiomatic circumstance where scarce minorities are relatively poor in their representations, we look for minority-enriched data distribution for the successful use of VRM.
Specifically, biased training mostly consisting of the majority not only leads to shifted boundaries towards the minority but also forms complex boundaries since majority groups become dominant in-between the minority groups~\cite{ibloss} as illustrated in Fig.~\ref{fig_embedding_illustration}~(Left).
To resolve these, we sequentially utilize two vicinal distributions formed by extrapolation and interpolation.
Extrapolation is to deal with distorted boundaries by enriching the minority whereas interpolation is to calibrate the biased boundaries.

Prior to explaining how we design these distributions, we introduce how they differ the behaviors between classes.
To be precise, we normalize the number of samples to be the tail-weighted criterion $\boldsymbol{\tau}$ for granting more weights on tail classes.
Given $\mathbf{q}$, the vector consisting of number of samples for all classes, the $s$-th component of vector $\boldsymbol{\tau}$ is defined as:
\begin{equation}
\label{eq_numsample}
    % \boldsymbol{\tau}_s = \frac{\mathbf{q}_s - \mathbf{q}_{\text{min}}}{\mathbf{q_{\text{max}}} - \mathbf{q}_{\text{min}}}
    \boldsymbol{\tau}_s = (\mathbf{q}_s - \mathbf{q}_{\text{min}}) / (\mathbf{q_{\text{max}}} - \mathbf{q}_{\text{min}}).
\end{equation}
$\mathbf{q}_s$, $\mathbf{q}_{\text{min}}$ and $\mathbf{q}_{\text{max}}$ stand for the $s$-th element, the minimum and maximum value in the vector $\mathbf{q}$, respectively.

% Prior to explaining how we design these distributions, we note how they differ the behaviors between classes.
% To be precise, we normalize the number of samples to be the tail-weighted criterion $\boldsymbol{\tau}$ for granting more weights on tail classes:
% \begin{equation}
% \label{eq_numsample}
%     % \boldsymbol{\tau}_s = \frac{\mathbf{q}_s - \mathbf{q}_{\text{min}}}{\mathbf{q_{\text{max}}} - \mathbf{q}_{\text{min}}}
%     \boldsymbol{\tau}_s = (\mathbf{q}_s - \mathbf{q}_{\text{min}}) / (\mathbf{q_{\text{max}}} - \mathbf{q}_{\text{min}})
% \end{equation}
% where $\boldsymbol{\tau}_s$ is the $s$-th component of vector $\boldsymbol{\tau}$ and $\mathbf{q}$ is the vector consisting of number of samples for all classes.
% $\mathbf{q}_{\text{min}}$ and $\mathbf{q}_{\text{max}}$ stand for minimum and maximum value in the vector $\mathbf{q}$, respectively.

% Prior to explaining how we adhere to the principle, we first introduce two vicinal distributions: interpolation and extrapolation.

% We achieve this by introducing two density estimates that adhere to the principle: the fewer the samples in each class, the more generated samples around that class distribution.
% Our TOMix has two purposes: extending the boundaries of minority samples towards the boundaries of majority samples and broadening the boundaries itself.
% To achieve these, we use two vicinity distributions, namely inter-interpolated and intra-extrapolated vicinities: 
% Dynamic 
% \subsubsection{Dynamic Extrapolation}
\textbf{Dynamic extrapolation.} 
Extrapolation estimates new data beyond the empirical distribution.
Although it is subject to generating samples with high uncertainty, we aim to generate probable data for tail classes to densify its vicinal distribution.
This is particularly crucial for the minority since only the specific partial data points are available as shown in Fig.~\ref{fig_embedding_illustration}~(Left).
% Be aware that extrapolation is only considered between samples belonging to the same class since we are intuitively aware that extrapolated label space may not be consistent with the complex feature space.
% Be aware that we only implement extrapolation between samples belonging to the same class since we are intuitively aware that extrapolated label space may not be consistent with the complex feature space.
% This is because embeddings are much more complex than just discrete label representation.
% To satisfy our goal, we use extrapolation between the instances belonging to the same class as described Eq.~\ref{eq:vicinity_extra}.

\begingroup
\setlength{\tabcolsep}{4.5pt} % Default value: 6pt
\renewcommand{\arraystretch}{0.9} % Default value: 1
\begin{table*}[t]
\footnotesize
\center
\begin{tabular}{l|c|cccccc|cccccc}
\hlineB{2.5}
\multicolumn{2}{c|}{} &\multicolumn{6}{c|}{ResNet-50} & \multicolumn{6}{c}{ResNet-101}\\ \hline
% \cite{cao2019learning, EQL,kang2019decoupling,cui2019class, mixup, videolt}.
LT-Methods & Agg & All & H& M& T& A@1 & A@5 & All& H & M & T & A@1& A@5 \\ \hlineB{2.5}
Baseline& \xmark& {0.499} & {0.675} & {0.553} & {0.376} & {0.650} & 0.828& {0.516}& {0.687}& {0.568}& {0.396}& {0.663}& 0.837 \\ 
LDAM~\cite{cao2019learning} &\xmark& {0.502} & {0.680} & {0.557} & {0.378} & {0.656} & 0.811& {0.518}& {0.687}& {0.572}& {0.397}& {0.664}& 0.820 \\ %\hline
EQL~\cite{EQL} &\xmark& {0.502} & {0.679} & {0.557} & {0.378} & {0.653} & 0.829& {0.518}& {0.690}& {0.571}& {0.398}& {0.664}& 0.838 \\ %\hline
CBS~\cite{kang2019decoupling} &\xmark& {0.491} & {0.649} & {0.545} & {0.371} & {0.640} & 0.820& {0.507}& {0.660}& {0.559}& {0.390}& {0.652}& 0.828 \\ %\hline
CB Loss~\cite{cui2019class} &\xmark& {0.495} & {0.653} & {0.546} & {0.381} & {0.643} & 0.823& {0.511}& {0.665}& {0.561}& {0.398}& {0.656}& 0.832 \\ %\hline
Mixup~\cite{mixup} &\xmark& {0.484} & {0.649} & {0.535} & {0.368} & {0.633} & 0.818& {0.495}& {0.660}& {0.546}& {0.381}& {0.641}& 0.843 \\ %\hline
Framestack~\cite{videolt}&\xmark& {0.516} & {0.683} & {0.569} & {0.397} & {0.658} & 0.834& {0.532}& {0.695}& {0.584}& {0.417}& {0.667}& 0.843 \\ \hline
Baseline &o&  0.676 & 0.793 & 0.717 & 0.585 & 0.702 & 0.861 & 0.690 & 0.806 & 0.730 & 0.601 & 0.715 & 0.869 \\ 
LDAM~\cite{cao2019learning} &o&  0.658 & 0.778 & 0.701 & 0.565 & 0.697 & 0.847 &	0.672 & 0.786 & 0.713 & 0.581 & 0.709 & 0.856 \\
EQL~\cite{EQL}  &o& 0.676 & 0.794 & 0.717 & 0.585 & 0.704 & 0.862 & 0.690 & 0.805 & 0.729 & 0.602 & 0.714 & 0.870 \\
CBS~\cite{kang2019decoupling} &o&  0.672 & 0.781 & 0.713 & 0.582 & 0.697 & 0.856 & 0.685 & 0.794 & 0.723 & 0.601 & 0.709 & 0.864 \\
CB Loss~\cite{cui2019class} &o& 0.681 & 0.784 & 0.717 & 0.601 & 0.695 & 0.847 & 0.693 & 0.793 & 0.728 & 0.616 & 0.705 & 0.854 \\
Mixup~\cite{mixup} &o&  0.686 & 0.795 & 0.726 & 0.598 & 0.703 & 0.861 & 0.699 & 0.807 & 0.737 & 0.614 & 0.715 & 0.867 \\ 
Framestack~\cite{videolt}  &o& 0.680 & 0.792 & 0.720 & 0.593 & 0.708 & 0.865 & 0.693 & 0.802 & 0.730 & 0.611 & 0.718 & 0.872 \\ \hline
\textbf{Ours} &o& \textbf{0.705} & \textbf{0.804} & \textbf{0.742} & \textbf{0.626} & \textbf{0.719} & \textbf{0.875}
& \textbf{0.719} & \textbf{0.815} & \textbf{0.753} & \textbf{0.644} & \textbf{0.730} & \textbf{0.883} \\ \hlineB{2.5}
\end{tabular}
% \vspace{-0.1cm}
\caption{Performance comparisons on VideoLT between long-tailed methods with features from ImageNet-pretrained ResNet-50 and ResNet-101. A@1, 5 are Top-1 and Top-5 accuracy and Agg indicates whether our learnable aggregators are applied or not.}
\label{Tab.Imagenet}
% \vspace{-0.4cm}
\end{table*}
\endgroup
Since our objective is to diversify with respect to sample numbers for each class, the sampling process for the extrapolation takes the number of samples into account.
% In short, the more samples each class contains, the more similar candidates are used for the extrapolation.
% This maintains the given data distribution as its vicinity for head classes since the output of extrapolation would be similar to given data points.
% On the other hand, different candidates are used for extrapolation for the minority to expand its vicinity as in Fig.~\ref{fig_embedding_illustration} (Center).
%We implement this by allowing extrapolation in-between different representations of the same instance where the difference between candidates are controlled by dynamic frame sampler.
We implement this by allowing extrapolation between different representations of the same instance where such difference comes from the Dynamic Frame Sampler~(DFS).
With the frame index set $\mathbf{T}=\{1, ..., T\}$ where $\left\vert \mathbf{T} \right\vert =T$, DFS generates a $t$-th frame of $s$-th class binary mask vector $\mathbf{m}$ as: % \in (0, 1)^{T}$ is derived as:
\begin{equation}
\label{eq_mask}
    \mathbf{m}_{s,t} = \mathds{1}_{[t \in \mathbf{I}_s]}, \\
\end{equation}
where $\mathds{1}$ is an indicator function that assigns 1 if $t\in \mathbf{I}_s$ and 0 otherwise. $\mathbf{I}_s$ is a disposable index set for $s$-th class that it is randomly generated at every computation defined as:
\begin{equation}
\label{eq_I}
    \mathbf{I}_s \subseteq \mathbf{T}, \;
    \left\vert \mathbf{I}_s \right\vert \sim  \textrm{U}(\text{max}(\floor*{\boldsymbol{\tau}_s \times T},\sigma), T), \; 
\end{equation}
% \begin{equation}
% \label{eq_mask}
% \begin{split}
%     % \mathbf{m}\rightarrow{}\{0, 1\}
%     & \mathbf{m}_{s,t} = \mathbbm{1}_{[t \in \mathbf{I}_s]}, \\
%     % \mathbf{m}_t = \begin{cases}
%     % 1, & \mbox{if }t\in \mathbf{I} \\
%     % 0, & \mbox{otherwise} \\
%     % \end{cases}, \\
%     & \mathbf{I}_s \subseteq \mathbf{T}, \;
%     \left\vert \mathbf{I}_s \right\vert \sim  \textrm{U}(\text{max}(\floor*{\boldsymbol{\tau}_s \times T},\sigma), T), \; 
%     \end{split}
% \end{equation}
where $\textrm{U}$ denotes an uniform distribution and $\sigma$ is the minimum value for $\left\vert\mathbf{I}_s\right\vert$ to prevent generating video prototypes only with the backgrounds.
Note that $\mathbf{m}_s$ is also resampled for every computation as $\mathbf{I}_s$ is disposable.
As the $s$-th class has more samples, $\left\vert\mathbf{I}_s\right\vert$ would be bigger in which more similar candidates are used for the extrapolation, thereby not many variations in generating $\mathbf{m}_s$.
For head classes, this maintains the given data distribution as its vicinity since the output of extrapolation would be similar to the given data points.
On the other hand, different candidates are used for extrapolation for the minority to expand its vicinity as varying $\mathbf{I}_s$ and $\mathbf{m}_s$ can be sampled.
% On the other hand, different candidates are used for extrapolation for the minority to expand its vicinity as in Fig.~\ref{fig_embedding_illustration} (Center).
Specifically, $\mathbf{m}_s$ is applied during the feature aggregation process.
For self-attentive aggregator, $\mathbf{m}_s$ is element-wisely multiplied to the input $\mathbf{x}^i$.
In opposition, for codebook-attentive aggregator, $\mathbf{m}_s$ is applied to Eq.~\ref{eq_vlad} as:
% Then, to apply $\mathbf{m}_s$ to the inputs for each aggregator, $\mathbf{m}_s$ is simply element-wise multiplied to the input of self-attentive aggregator. 
% On the other hand, $\mathbf{m}_s$ is applied to Eq.~\ref{eq_vlad} as:
\begin{equation}
\label{eq_masking}
% \vspace{-0.3cm}
        \mathbf{\tilde{x}}_{c,k}^{i} = \sum^{T}_{t=1}\mathbf{m}_{\mathbf{y}^{i},t}\:\rho_k(\mathbf{x}_{t}^{i})(\mathbf{x}^{i}_{t,c}-\boldsymbol{\mu}^{i}_{k,c}),
        % &D(i,k) = D(i, k) \cdot \mathbf{m}_{i},\;\; \\
    % \end{split}
    % \vspace{-0.1cm}
\end{equation}
to mask out the specific instance.
% Eq.~\ref{eq_zglobal} and \ref{eq_vlad} as:
% \vspace{-0.05cm}
% \begin{align}
% \label{eq_masking}
%     % \begin{split}
%         &\mathbf{x}_{global}^{i} = \text{Agg}(\mathbf{x}^{i} \circ \mathbf{m}_{\mathbf{y}^{i}}),\;\; \mathbf{x}_{local}^{i} = \mathbf{x}^{i} \circ \mathbf{m}_{\mathbf{y}^{i}},\\[-2.5pt]
%         % & z_{local} = z \cdot \mathbf{m}^{l},\; \mathbf{m}^l \in (0, 1)^T \\
%         & \mathbf{\tilde{x}}_{c,k}^{i} = \sum^{T}_{t=1}\mathbf{m}_{\mathbf{y}^{i},t}\:\rho_k(\mathbf{x}_{t}^{i})(\mathbf{x}^{i}_{t,c}-\boldsymbol{\mu}^{i}_{k,c}),
%         % &D(i,k) = D(i, k) \cdot \mathbf{m}_{i},\;\; \\
%     % \end{split}
%     \vspace{-0.3cm}
% \end{align}
% taking the label $\mathbf{y}_i$ into account, and $\circ$ denotes element-wise multiplication.
% so that the minority with smaller $\tau_{s}$ has a high probability of being represented differently every time it is aggregated.
% Since $\mathbf{m}$ has higher probability to have more 1s for the head classes with high $\boldsymbol{\tau}^s$, the possible expression would be limited for the same instance.
% On the other hand, t
This encourages aggregators to diversify the tail classes because it is highly likely to express the minority differently whenever features of the minority are aggregated.
Then, $p_{ex}$, the minority-diversified vicinal distribution is created as varying cases of the cartesian product of possible prototypes from aggregators are given as input to the extrapolation process as:
% Then, $p_{ex}$, the vicinal distribution with minority-oriented feature space is created as varying cases of the cartesian product of possible prototypes from two aggregators are given as input to the extrapolation process as:
\begin{equation}
\label{eq:vicinity_extra}
% \resizebox{0.90\hsize}{!}{$%
     p_{ex}(\hat{\mathbf{z}}, \hat{\mathbf{y}} | \mathbf{u}^i, \mathbf{v}^{i}, \mathbf{y}^i) = \mathbb{E}_{\omega}\big[\delta(\hat{\mathbf{z}}=\omega \mathbf{u}^i + (1-\omega) \mathbf{v}^i, \hat{\mathbf{y}}= \mathbf{y}^i )\big], 
    %  \;\omega \sim \text{Beta}(\alpha, \alpha) + 1,
    %  \vspace{-0.05cm}
% v_{ex}(\hat{\mathbf{z}}, \hat{\mathbf{y}} | \mathbf{z}_{i}, \mathbf{y}_{i}) = \frac{1}{N}\sum_{j=1}^{N}\mathbbm{1}_{[\mathbf{y}_{i} = \mathbf{y}_{j}]}\mathbb{E}_{\omega}\big[\delta(\hat{\mathbf{z}}=\omega \mathbf{z}_i + (1-\omega) \mathbf{z}_j, \hat{\mathbf{y}}= \mathbf{y}_i )\big], \;\omega \in [1, 2].
\end{equation}
where $\mathbf{u}^i$ and $\mathbf{v}^i$ stand for two different prototypes from instance $\mathbf{x}^i$.
$\omega$ is sampled as $\omega \sim \text{Beta}(\alpha, \alpha) + 1$, bounded between [1,~2] to prevent the novel data from deviating much from class boundaries.
% For the mixing ratio, we simply define $\omega \sim \text{Beta}(\alpha, \alpha) + 1$ to prevent the novel data from deviating much from class boundaries since $\omega$ is bounded between [1, 2].
$\delta$ denotes Dirac mass function.
Accordingly, the fewer the samples each class has, the larger the vicinal regions considered as a possible data distribution that smooths the boundaries as shown in Fig.~\ref{fig_embedding_illustration} (Center).
Note that extrapolation is only considered between the prototypes of the same instance for stability.

% In short, the difference gap in the candidates according to class frequency, leads to expanded vicinity only for the minority as shown in Fig.~\ref{fig_embedding_illustration} (Center).

\textbf{Calibrated interpolation. }
% \subsubsection{Calibrated Interpolation}
% Data interpolation, also known as Mixup~(\cite{mixup}, \cite{manifoldmixup}), generates a pair of data-label by the weighted sum of the given data and one-hot labels.
Whereas the extrapolation is to smooth the distorted boundaries, vicinal distributions of sparse instances may not cover the large true distribution. 
Thus, we conduct interpolation, also known as Mixup~\cite{mixup, manifoldmixup}, to balance the shifted boundaries in-between the outputs from extrapolation.
% Thus, we implement interpolation with the output of extrapolation, also known as Mixup~\cite{mixup, manifoldmixup}, to balance the shifted boundaries.
Interpolation makes it simple by modeling the linear relation between different classes.
Still, the naive use of interpolation has been shown to have not much benefit on VLTR, where drastic gaps exist between the sample numbers for classes~\cite{videolt}.
This is because the difference between the class distributions was not taken into account when assigning the vicinity relation across the classes.
% Whereas the extrapolation is to smooth the distorted boundaries, interpolation, also known as Mixup~(\cite{mixup}, \cite{manifoldmixup}), is to balance the shifted boundaries.
% This is because vicinal distributions of sparse data points may not cover the large true distribution.
% Interpolation makes it simple by modeling the linear relation between different classes.
% Still, the naive use of interpolation has been shown to have no benefits on VLTR where drastic gaps exist between the sample numbers for classes~(\cite{videolt}).
% Briefly, we use interpolated distribution, namely mixup~(\cite{mixup}), for its simplicity to train the linear behavior of the model even though it is shown to have no benefits on VLTR where drastic gaps exist between the number of samples for each class.

With minority-densified pair $(\mathbf{\hat{z}}^{i}, \mathbf{\hat{y}}^{i})$, we thus, adjust biased training and balance the boundaries by enforcing the minority classes to occupy larger regions in the linearly interpolated vicinal space.
To be specific, we grant more weights on the minority's label space by multiplying the smoothing factor derived from $\boldsymbol{\tau}$.
With the calibrated label space, vicinal distribution $p_{in}$ can be defined as:
% 인터폴레이션으로 인한 vicinity distribution은 다음과같이 정의된다:
% \begin{equation}
% \label{eq:vicinity_inter}
% \begin{split}
%      p_{in}(\tilde{\mathbf{z}}, \tilde{\mathbf{y}} | &(\mathbf{\hat{z}}^{i}, \hat{\mathbf{y}}^{i}) \sim p_{ex}) =  \frac{1}{N}\sum_{j=1}^{N}\mathbb{E}_{\lambda}\big[\delta(\tilde{\mathbf{z}}=\lambda \mathbf{\hat{z}}^i + (1-\lambda) \mathbf{\hat{z}}^j, \\
%      & \tilde{\mathbf{y}}_s = \min_{1 \le s \le S}(\big(\lambda \hat{\mathbf{y}}^i_s + (1-\lambda) \hat{\mathbf{y}}^j_s\big) \times ((1-\boldsymbol{\tau}_s) + \gamma), 1)\big], \; \lambda \sim \text{Beta}(\alpha, \alpha)
%     %  \hat{y}=\lambda y_i + (1-\lambda) y_j)\big], \;\lambda \in [0, 1]\\
% \end{split}
% \end{equation}
\begingroup
\thickmuskip=0.3\thickmuskip
\begin{align}
\label{eq:vicinity_inter}
\begin{split}
     p_{in}&(\tilde{\mathbf{z}}, \tilde{\mathbf{y}} | (\mathbf{\hat{z}}^{i}, \hat{\mathbf{y}}^{i}) \sim p_{ex}) =  \frac{1}{N}\sum_{j=1}^{N}\mathbb{E}_{\lambda}\big[\delta(\tilde{\mathbf{z}}=\lambda \mathbf{\hat{z}}^i + (1-\lambda) \mathbf{\hat{z}}^j, \\[-2pt]
     & \tilde{\mathbf{y}}_s = \min_{1 \le s \le S}(\big(\lambda \hat{\mathbf{y}}^i_s + (1-\lambda) \hat{\mathbf{y}}^j_s\big) \times ((1-\boldsymbol{\tau}_s) + \gamma), 1)\big], 
    %  \; \lambda \sim \text{Beta}(\alpha, \alpha)
    %  \hat{y}=\lambda y_i + (1-\lambda) y_j)\big], \;\lambda \in [0, 1]\\
\end{split}
\end{align}
\endgroup
where $1-\boldsymbol{\tau}_s$ makes the minority to have bigger weights. 
$\gamma$ and $\lambda$ are the smoothing bias to prevent from making the label to zero and the mixing ratio, respectively.
$\lambda$ is defined to be sampled as $\lambda \sim \text{Beta}(\alpha, \alpha)$.
$S$ and $\mathbf{y}^i_s$ each denote the number of classes and $s$-th class label for $i$-th instance.
Also, despite the low chance of mixing the same tail class instances, we bound the upper value of one-hot label to 1.
% $\lambda$ is a mixing ratio sampled from beta distribution, $\text{Beta}(\alpha, \alpha)$.
Similar to the vicinity formed by Eq.~\ref{eq:vicinity_extra}, calibrated interpolation directly reflects the class frequency by allocating the minority more space between distributions of different classes than the majority.
With all parts integrated, optimization in Eq.~\ref{eq_vrm} is implemented with pairs of $\tilde{\mathbf{z}}^i$ and $\tilde{\mathbf{y}}^i$ sampled from $p_{in}$.

%% file: 4_experiment.tex
\section{Experiments}

\begingroup
\setlength{\tabcolsep}{4pt} % Default value: 6pt
\renewcommand{\arraystretch}{0.9} % Default value: 1
\begin{table}[!t]
    \centering
    \footnotesize
    % \vspace{-0.1cm}
    % \vspace{-0.05cm}
    \begin{tabular}{c|c|cccccc}
        \hlineB{2.5}
        Methods & Agg &\multicolumn{1}{c}{All} & H & M & T & A@1 & A@5 \\ \hlineB{2.5} %
         % \cline{1-1} \cline{3-6} \cline{8-11} 
        % baseline & 0.565 & 0.757 & 0.620 & 0.436  \\
        % LDAM & 0.565 & 0.750 & 0.620 & 0.439\\
        % EQL& 0.567 & 0.757 & 0.623 & 0.439 \\
        % CBS& 0.558 & 0.733 & 0.612 & 0.435  \\
        % CB Loss& 0.563 & 0.744 & 0.616 & 0.440\\
        % Mixup& 0.548 & 0.736 & 0.602 & 0.425 \\
        % Framestack & 0.580 & 0.759 & 0.632 & 0.459 \\ \hlineB{2.5}% \cline{1-1}
        Baseline &\xmark & 0.565 & 	0.757 & 0.620 &	0.436 & - &-\\
        Fr.stack &\xmark & 0.580 & 0.759 & 0.632 & 0.459 & -&-\\
        \hline
        Baseline &o&0.680 & 	0.825 & 	0.727 & 	0.575 & 0.714 & 0.873\\
        LDAM &o& 0.667 & 	0.807 & 	0.712 & 	0.566 & 0.708 & 0.863\\
        EQL &o& 0.680 & 	0.825 & 	0.727 & 	0.575  & 0.713 & 0.874 \\
        CBS &o& 0.674 & 	0.816 & 	0.721 & 	0.570  & 0.709 & 0.869\\
        CB Loss &o& 0.681 & 	0.818 & 	0.724 & 	0.584  & 0.706 & 0.862\\
        Mixup &o& 0.688 & 	0.828 & 	0.733 & 	0.588 & 0.714 & 0.874\\
        Fr.stack &o& 0.688 & 	0.825 & 	0.732 & 	0.590 & 0.716 & 0.879\\ \hline
        % \cline{3-6} \cline{8-11} 
        \textbf{Ours} &o& \textbf{0.704} & \textbf{0.833} & \textbf{0.746} & \textbf{0.610}  & \textbf{0.725} & \textbf{0.885}  \\ \hlineB{2.5}
    \end{tabular}
    % \vspace{-0.1cm}
    \caption{Results on VideoLT with features from TSM.}
    \label{Tab.Kinetics}
    % \vspace{-0.4cm}
\end{table}
\endgroup

\subsection{Evaluation settings and datasets}
\textbf{Evaluation settings.} 
% We follow Framestack~\cite{videolt} to use both the Image- and Video-pretrained networks to extract features; ImageNet-pretrained ResNet-50, 101~\cite{imagenet, resnet} and Kinetics-pretrained TSM~\cite{kinetics-400, tsm} using ResNet-50 are employed.
% Average precision (AP) and accuracy are used for the metric.
% In all tables throughout this section, we abbreviate head, medium, and tail to H, M, and T, and the best results are in bold.
For the evaluation, we follow the settings from Framestack~\cite{videolt} as we use both Image-pretrained and Video-pretrained networks to extract features from the penultimate layer; ImageNet-pretrained ResNet-50, 101~\cite{imagenet, resnet} and Kinetics-pretrained TSM~\cite{kinetics-400, tsm} using ResNet-50 are employed.
Average precision (AP) and accuracy are used for the metric.
For more details about implementation details and settings, we refer to the appendix and the implementation~\footnote{https://github.com/wjun0830/MOVE}.
In all tables throughout this section, we abbreviate head, medium, and tail to H, M, and T, and the best results are in bold.

% For the metrics, we use two popular metrics for video recognition tasks: average precision (AP) and accuracy.
% In all tables throughout this section, we abbreviate head, medium, and tail to H, M, and T, and the best results are in bolds.

% \textbf{Evaluation settings.} To validate the generalization capability of our approach for VLTR, we follow experimental settings from Framestack~\cite{videolt}.
% We use both Image-pretrained and Video-pretrained networks to extract features from the penultimate layer. 
% Specifically, ImageNet-pretrained ResNet-50, 101~\cite{imagenet, resnet} and Kinetics-pretrained TSM~\cite{kinetics-400, tsm} using ResNet-50 are employed.
% For the evaluation, we use two popular metrics for video recognition tasks: average precision (AP) and accuracy.
% In long-tailed recognition, it is important to improve the model's performance on tail classes without sacrificing head classes.
% Hence, we show the performance gap regarding the data distribution by measuring AP for each group of the head, medium, and tail classes.
% In all tables throughout this section, we abbreviate head, medium, and tail to H, M, and T, and the best results are in bolds.

\begingroup
\setlength{\tabcolsep}{4.5pt} % Default value: 6pt
\renewcommand{\arraystretch}{0.9} % Default value: 1
\begin{table*}[!t]
\scriptsize
\center
\begin{tabular}{c|c|cc | cc | cc | cc | cc | cc | cc | cc}
\hlineB{2.5}
\multicolumn{2}{c|}{} &\multicolumn{8}{c|}{ResNet-50} & \multicolumn{8}{c}{ResNet-101}\\ \hline
\multicolumn{2}{c|}{Imbalance Ratio} & \multicolumn{2}{c|}{0.01} & \multicolumn{2}{c|}{0.02} & \multicolumn{2}{c|}{0.05} & \multicolumn{2}{c|}{0.1} & \multicolumn{2}{c|}{0.01} & \multicolumn{2}{c|}{0.02} & \multicolumn{2}{c|}{0.05} & \multicolumn{2}{c}{0.1} \\ \hline
LT-Methods & Agg & AP & ACC & AP & ACC & AP & ACC & AP & ACC & AP & ACC & AP & ACC & AP & ACC & AP & ACC  \\ \hlineB{2.5}
Baseline & \xmark & 0.466 & 0.397 & 0.510 & 0.456 & 0.555 & 0.525 & 0.589 & 0.573 & 0.492 & 	0.429 & 0.534 & 0.480 & 0.579 & 0.548 & 0.611 & 0.594 \\
Framestack & \xmark & 0.477 & 0.410 & 0.518 & 0.465 & 0.560 & 0.531 & 0.594 & 0.579 & 0.504 & 0.438 & 0.543 & 0.490 & 0.581 & 0.557 & 0.615 & 0.596 \\ \hline
Baseline & o & 0.559 & 0.490 & 0.595 & 0.531 & 0.633 & 0.591 & 0.662 & 0.623 &  0.581 & 0.511 & 0.619 & 0.562 & 0.658 & 0.611 & 0.686 & 0.640 \\
CB Loss & o & 0.549 & 0.440 & 0.593 & 0.497 & 0.633 & 0.565 & 0.666 & 0.613 &   0.573 & 0.470 & 0.615 & 0.521 & 0.655 & 0.584 & 0.688 & 0.624 \\
Mixup & o & \textbf{0.570} & 0.488 & 0.608 & 0.533 & 0.643 & 0.588 & 0.672 & 0.626 &	    0.592 & 0.499 & 0.629 & 0.547 & 0.666 & 0.602 & 0.694 & 0.637 \\
Framestack & o & 0.556 & 0.480 & 0.596 & 0.525 & 0.631 & 0.582 & 0.664 & 0.620 &    0.574 & 0.499 & 0.616 & 0.545 & 0.652 & 0.598 & 0.682 & 0.634 \\ \hline
\textbf{Ours} & o & \textbf{0.570} & \textbf{0.509} & \textbf{0.609} & \textbf{0.553} & \textbf{0.646} & \textbf{0.604} & \textbf{0.675} & \textbf{0.636} & \textbf{0.593} & \textbf{0.528} & \textbf{0.632} & \textbf{0.577} & \textbf{0.667} & \textbf{0.626} & \textbf{0.697} & \textbf{0.655} \\ \hlineB{2.5}
\end{tabular}
% \vspace{-0.1cm}
\caption{Performances comparison on varying scenarios with Imbalanced-MiniKinetics200.}
% \vspace{-0.4cm}
\label{Tab.Minikinetics}
\end{table*}
\endgroup

\begingroup
\setlength{\tabcolsep}{2.4pt} % Default value: 6pt
\renewcommand{\arraystretch}{0.9} % Default value: 1
\begin{table}[t!]
    \centering
    \footnotesize
    \begin{tabular}{c|cc|cc|cccccc}
        \hlineB{2.5}
        \multicolumn{1}{c|}{} & \multicolumn{2}{c|}{Agg} & \multicolumn{2}{c|}{MOVE}& \multicolumn{6}{c}{ResNet-101} \\ \hline
         & {S-A} & {C-A} & Ex. & In & {All}& {H} & {M} & {T} & {A@1}& A@5\\ \hlineB{2.5}
        (a) & {}  & {}  &    & & {0.516}& {0.687}& {0.568}& {0.396}& {0.663}& 0.837\\ \hline
        (b)&{\cmark} & {}  & &  & {0.681}& {0.795}& {0.719}& {0.595}& {0.706}& 0.866\\ \hline
        (c) & & \cmark  &   & & 0.677 & 0.787 & 0.714 & 0.593 & 0.716 & 0.863 \\ \hline
        (d) & {\cmark} & {\cmark}& & & 0.690 & 0.806 & 0.730 & 0.601 & 0.715 & 0.869 \\ \hline
        (e)& \cmark & \cmark &  \cmark &  & 0.710 & \textbf{0.817} & 0.746 & 0.632 & 0.725 & 0.878 \\ \hline
        (f)& \cmark & \cmark &  & \cmark & 0.712 & 0.810 & 0.749 & 0.632 & 0.726 & 0.876 \\ \hline
        % \cmark & \cmark &  \cmark &  & 0.715 & 0.815 & 0.751 & 0.636 & 0.729 & 0.883 \\ \hline
        (g)& \cmark & \cmark & \cmark & \cmark & \textbf{0.719} & 0.815 & \textbf{0.753} & \textbf{0.644} & \textbf{0.730} & \textbf{0.883}\\ \hlineB{2.5}
        % \cmark & \cmark & \cmark && \cmark & 0.719 & 0.815 & 0.753 & 0.644 & 0.730 & 0.883\\
    \end{tabular}
    % \vspace{-0.1cm}
    \caption{Ablation study. S-A and C-A are self-attentive and codebook-attentive aggregators, and Ex. and In. are abbreviated keywords for dynamic extrapolation and calibrated interpolation. 
    ResNet-101 features are used.}
    \label{Tab.Ablation1}
    % \vspace{-0.3cm}
\end{table}
\endgroup

\noindent\textbf{Datasets.} 
% Additional to large-scale VideoLT, we introduce Imbalanced-MiniKinetics200, synthetically created, thereby more flexible to evaluate various imbalanced cases.
Additional to large-scale VideoLT dataset which consists of 256218 videos of 1004 classes, we synthetically create and introduce Imbalanced-MiniKinetics200 for VLTR.
Imbalanced-MiniKinetics200 is more flexible dataset that can be manipulated to evaluate various imbalanced cases.
Specifically, we sample up to 400 videos for each of the 200 classes and provide settings to simply manipulate the data distribution by controlling the value of the imbalanced ratio~\cite{xie2017rethinking, cui2019class}.
This makes it more challenging since the minority does not contain sufficient samples to be trained on.
Details are in the appendix.
% Furthermore, to test on varying imbalanced scenarios, we create Imbalanced-MiniKinetics200 with varying imbalance ratios~\cite{xie2017rethinking, cui2019class}.
% Details are in the appendix.

% \noindent\textbf{Datasets.} VideoLT and Imbalanced-MiniKinetics200 are used for evaluation. 
% VideoLT is a large-scale dataset that consists of 256,218 videos of 1004 classes. 
% Furthermore, to test on varying imbalanced scenarios, we create Imbalanced-MiniKinetics200 with varying imbalance ratios~\cite{xie2017rethinking, cui2019class}.
% Details are in the appendix.

% Specifically, we sample up to 400 videos for each of 200 classes from Kinetics dataset~\cite{kinetics-400, xie2017rethinking}. 
% Then, we used imbalance ratio $\mu$ to manipulate the harshness in imbalanced setting~\cite{cui2019class, cao2019learning} by generating a imbalance vector $v \in (\mu, 1)^K$ where K denotes the number of classes.
% Then, $v$ is multiplied class-wise as $n_i = n_{max}v_i$ where $n_i$ and $n_{max}$ indicate number of samples for i-th class and maximum number of videos in certain class.

\begin{figure}[t]
    \centering
    \includegraphics[width=0.45\textwidth]{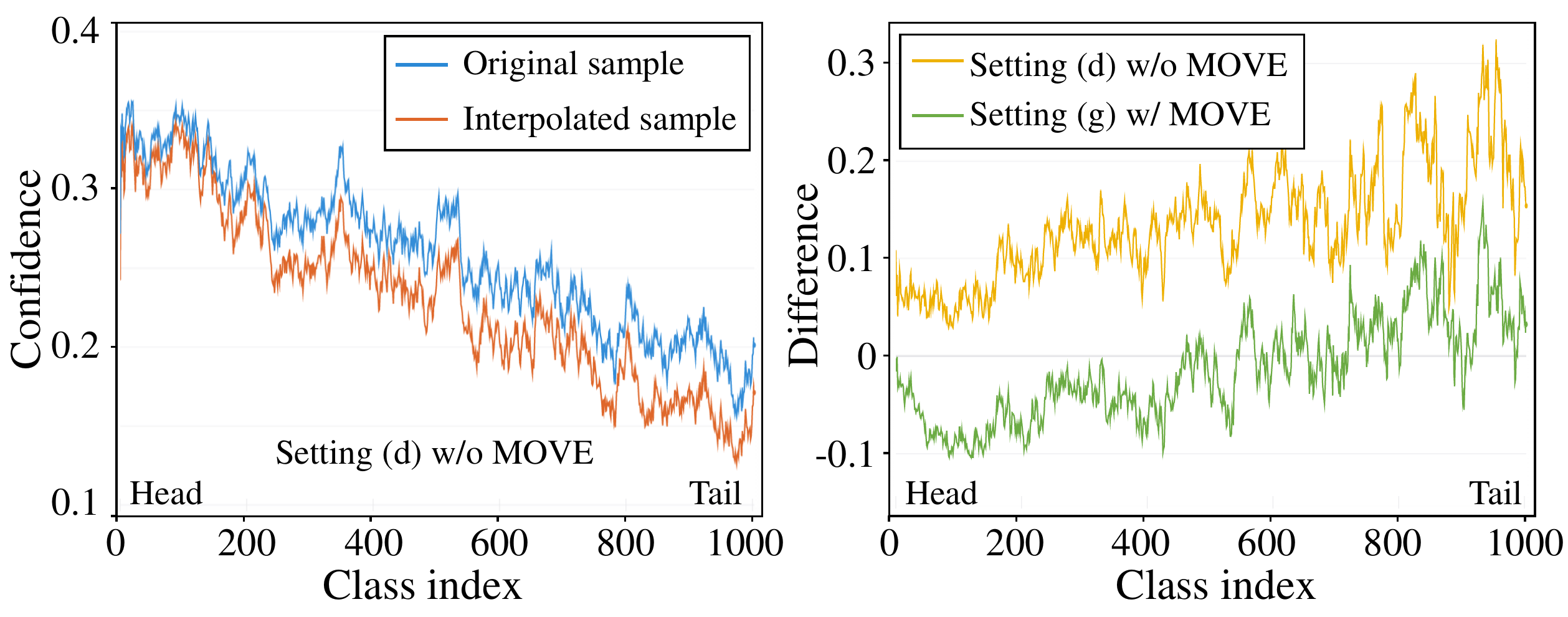}
    % \vspace{-0.2cm}
    \caption{
    Biased and distorted decision boundaries.
    (\textbf{Left}) Per-class average confidence for original and interpolated samples for setting (d) in Tab.~\ref{Tab.Ablation1}.
    Interpolation is performed within the same class.
    The confidence gap between head and tail classes implies biased boundaries. 
    (\textbf{Right}) Normalized difference between the confidence of original and interpolated samples for settings (d), (g).
    Normalized difference in yellow line for (d) is computed with blue and orange vectors in the left figure as $\text{yellow} = \frac{\text{blue} - \text{orange}}{\text{blue}}$.
    For setting (d), the yellow line, increasing difference towards the tail classes implies the distorted boundaries since the interpolated samples within the class are less likely to be class-relevant.
    On the other hand, our proposed MOVE lowers the difference across all classes.
    The decreased difference in tail classes reveals the smoothed boundaries since interpolated samples are likely to reside within the class boundary.
    Regarding the decline in head classes, we think that the model's learned linearity elicited the phenomenon since interpolation within the class results in the highest density at the class center, where the highest confidence scores should be present.
    }
    \label{fig_graph}
    % \vspace{-0.3cm}
\end{figure}

\subsection{Comparison with state-of-the-art}
\textbf{VideoLT} We compare MOVE against previous methods for long-tailed recognition: LDAM+DRW, EQL, CBS, CBLoss, Mixup, and Framestack in Tab.~\ref{Tab.Imagenet}.
%~\cite{cao2019learning, EQL,kang2019decoupling,cui2019class, mixup, videolt}.
With representations from ImageNet-pretrained backbones, our method outperforms previous methods without our aggregators by a large margin, establishing new state-of-the-art AP and accuracy scores.
To our superior performance over baselines, we state that this is because VLTR poses more difficulties compared to the image domain. 
% The success of our method comes from the consideration of additional problems: using task-irrelevant features due to computational limitations to training the whole network and only the video-level label available.
The success of our method come from two components.
First, we considered two crucical problems in VLTR by addressing task-irrelevant features without excessively training the whole network and bridging the gap in the level of supervision and video data.
To verify these, we also test the applicability of aggregators to the baselines. As reported with Agg, our aggregators consistently provide significant performance boosts even for the baselines. Those results confirm that the merits of aggregators that produce task-relevant features from weakly-labeled video data, and re-establishes finetuned baselines. 
Second, MOVE alleviates the biased training and assist the model to generalize on imbalanced settings.
As a result, it is shown that MOVE clearly provides higher performances than previous techniques even if our aggregators are attached, especially for tail classes. 

We also show the superiority of our approach with extracted features from Kinetics-pretrained TSM in Tab.~\ref{Tab.Kinetics}.
Unlike the ImageNet-pretrained backbones, video-pretrained backbones are capable of maintaining the temporal consistency in the video.
% Generally, temporal consistency improves the quality of frame level features which boosts the performance of our baselines, whereas there is no additive benefits on our approach since we aggregate features at the video level.
% Thus, our method is robust to existence of temporal consistency of the pretrained network.
Generally, it improves the quality of frame-level features, achieving performance gains for baselines.
Yet, the gains are less compared to using our aggregators.
% Generally, it improves the quality of frame-level features, which boosts the performance of the baselines even though there are less benefits to approaches with aggregators.
% since we aggregate representations at the video level.
This is because our aggregators discover self~(i.e., temporal) and global relationships to manipulate the class-discriminative features in which the temporal relations are already taken into account as their sub-objective.
Likewise, it is shown that employing our aggregators is much more beneficial than deploying video pretrained network in Tab.~\ref{Tab.Imagenet} and Tab.~\ref{Tab.Kinetics}.
In addition, MOVE outperforms all long-tailed approaches with noticeable margin especially on tail classes.

% In Tab.~\ref{Tab.Minikinetics}, we find that MOVE outperforms baselines in all scenarios even when our learnable aggregators are applied.

\textbf{Imbalanced-MiniKinetics200} For further validations in various imbalanced scenarios, we synthetically create Imbalanced-MiniKinetics200 with varying imbalanced ratios. 
% In Tab.~\ref{Tab.Minikinetics}, we attach our proposed learnable aggregators to the baselines to fairly compare MOVE with other long-tailed recognition methods since aggregators are highly applicable to boost all baselines.
In Tab.~\ref{Tab.Minikinetics}, we compare new baselines~(with our aggregators) with MOVE since aggregators are highly applicable to boost all baselines.
% In Tab.~\ref{Tab.Minikinetics}, we use our proposed learnable aggregators as a new baseline and compare long-tailed recognition methods. 
Although Mixup achieves performance gains in terms of AP by its positive impacts on confidence calibration~\cite{zhong2021improving,thulasidasan2019mixup}, it has shown a limitation in balancing distributions thereby losing overall accuracy. Furthermore, Framestack especially struggles on a relatively small-scale imbalanced dataset. Likewise, whereas other long-tailed recognition approaches are less effective on Imbalanced-MiniKinetics200, it is observed that our proposed techniques, the use of aggregators and MOVE, have benefits across all the tested dataset sizes and harshness of the data distribution. Note that, Imbalanced-MiniKinetics200 provides more challenging testbeds since fewer numbers of videos per class are available for the minority. Therefore, we believe that our newly introduced dataset would be a strong standard benchmark along with VideoLT and will stimulate future research for VLTR.

\subsection{Ablation study}
% Sec.4.Ablation
\label{Sec.4.Ablation}
\textbf{Effectiveness of aggregators and MOVE.} 
To evaluate the importance of different components, we conducted an ablation study in Tab.~\ref{Tab.Ablation1}.
Specifically, we investigate the impact of self-attentive, codebook-attentive representations, and each component of MOVE.
We can see that both self-attentive and codebook-attentive representations resolve the domain gap and label uncertainty, resulting in improved overall performance~((b), (c), (d)).
Furthermore, MOVE clearly appears to be effective in long-tailed distribution by boosting performance on tail classes up to 5.2\% in (e) and (f) compared to (d).
We conjecture that improvements in head classes are from the calibrated boundaries leading the model to generalize on head classes with alleviated overfitting.
These results highlight the importance of calibrating decision boundaries for tail classes in VLTR.
All components combined in (g), we describe how it relieves the issue of biased and distorted boundaries compared to (d) in Fig.~\ref{fig_graph}.
% To evaluate the importance of different components of our approach, we conducted an ablation study in Tab.~\ref{Tab.Ablation1}, observing the change in performance.
% Specifically, we investigate the impact of self-attentive, codebook-attentive representations, and each component of MOVE.
% We can see that both self-attentive and codebook-attentive representations resolve the domain gap and label uncertainty, resulting in improved overall performance~((b), (c), (d)).
% Furthermore, MOVE clearly appears to be effective in long-tailed distribution by boosting performance on tail classes up to 5.2\% in (e) and (f) compared to (d).
% We conjecture that improvements in head classes are from the calibrated boundaries leading the model to generalize on head classes with alleviated overfitting.
% These results highlight the importance of calibrating decision boundaries for tail classes in VLTR.
% All components combined in (g), we describe how it relieves the issue of biased and distorted boundaries compared to (d) in Fig.~\ref{fig_graph}.

\begingroup
\setlength{\tabcolsep}{5pt} % Default value: 6pt
\renewcommand{\arraystretch}{0.72} % Default value: 1
\begin{table}[t]
	\footnotesize
	\centering
    \begin{tabular}{c|cccccc}
    \hlineB{2.5}
    LT Methods & \multicolumn{1}{c}{All} & H & M & T & A@1 & A@5 \\ \hlineB{2.5}
    baseline & 0.565 & 0.757 & 0.620 & 0.436 & - & - \\
    MSA & 0.642 & 0.794 & 0.689 & 0.535 & 0.685 & 0.848 \\
    PSA & \textbf{0.669} & \textbf{0.817} & \textbf{0.716} & \textbf{0.563} & \textbf{0.702} & \textbf{0.867}\\
    \hlineB{2.5}
    \end{tabular}
    % \vspace{-0.1cm}
    \caption{Comparisons between MSA and PSA for verifying the importance of resolving each challenge in VLTR.}
    % \vspace{-0.3cm}
    \label{Tab.Ablation2}
\end{table}
\endgroup
\textbf{Importance of PSA.}
To further split and verify the need of handling two additional challenges in VLTR,
we make a thorough analysis of self-attentive representations by comparing PSA with conventional MSA.
MSA, an attention module to yield task-relevant features, also works as temporal smoothing and grants temporal consistency between frames~\cite{park2022vision}.
% MSA is an attention module that yields task-relevant features while it also works as temporal smoothing and grants temporal consistency between frames~\cite{park2022vision}.
Hence, features from TSM are used to verify the effect of creating a task-relevant feature, excluding the impact of MSA being a smoothing module.
Note that, in Tab.~\ref{Tab.Ablation2}, MSA may partially resolve the label uncertainty issue since background scenes can also be re-expressed with class-relative features, potentially indicating the importance of two problems.
Nevertheless, we observe that video-prototypical representations clearly address the label uncertainty by matching the degree of levels between the feature and the label.

%% file: 5_related_work.tex
\section{Related Work}
% \subsection{Long-tailed Recognition}
%\textbf{Long-tailed Recognition}
Long-tailed recognition aims to overcome the data imbalance~\cite{kubat1997addressing, karakoulas1998optimizing, wang2017learning, kang2019decoupling, tang2020long} in which the collected data naturally tend to follow long-tailed class distributions in real-world applications.
Since they mainly assume that the performance decline in minority classes is caused by the biased decision boundary of the classifier~\cite{ibloss}, most of the recent studies have focused on debiasing the classifier~\cite{dong2017class, wang2020deep, nam2020learning, chu2020feature, yang2020rethinking, liu2020deep, zhu2021cross, zhang2021distribution}.
A popular stream is to re-sample.
For instance, over-sampling the minority~\cite{chawla2002smote, shen2016relay, buda2018systematic, remix, m2m, li2021metasaug, wang2022label} and under-sampling the majority~\cite{japkowicz2002class, more2016survey, buda2018systematic} to train the model with balanced class distribution have shown to be effective.
% For instance, over-sampling was proposed to train the model with balanced data distribution between classes~\cite{chawla2002smote, shen2016relay, buda2018systematic, remix, m2m, li2021metasaug, wang2022label} and there were also approaches using under-sampling strategies on majority classes~\cite{japkowicz2002class, more2016survey, buda2018systematic}.
% \cite{chawla2002smote, shen2016relay, buda2018systematic, remix} proposed to train the model with the minority over-sampling technique to balance the data distribution, and \cite{japkowicz2002class, more2016survey, buda2018systematic} instead employed under-sampling strategies on majority classes.
Other streams include re-weighting~\cite{huang2016learning, cao2019learning, cui2019class}, improving the quality of representation~\cite{moon2022tailoring, yang2020rethinking}, and adopting multi-experts~\cite{ride, tadelongtail}.
Although MOVE shares the general motivation with them, it only dynamically diversifies the prototype for tail classes while retaining the data frequency per class.
% Re-weighting approaches assign different weights to different classes considering the data frequency in each class~\cite{huang2016learning, cao2019learning, cui2019class} and the latter methods usually utilize self-supervised learning techniques~\cite{moon2022tailoring, yang2020rethinking}.
% Other streams include re-weighting and improving the quality of feature representation. 
% Re-weighting approaches assign different weights to different classes considering the data frequency in each class~\cite{huang2016learning, cao2019learning, cui2019class} and the latter methods usually utilize self-supervised learning techniques~\cite{moon2022tailoring, yang2020rethinking}.

% Although these methods have been effective at the image level, generalizing them to video-level applications is another challenge.
In addition, generalizing image-level approaches to video-level is another challenge.
For instance, the re-weighting scheme is known to interrupt model optimization on large datasets.~\cite{cui2019class}, whereas the goal of VLTR is on handling the vast amount of video.
The presence of temporal dimension and the fact that video labels are mainly upon the video-level supervision are other reasons why even renowned methods in the image domain are difficult to generalize in VLTR.
% Recently, Framestack~\cite{videolt} was proposed for VLTR which adaptively stacks different number of frames for different classes according to the model performance.
Recently, Framestack~\cite{videolt} was proposed for VLTR which adaptively stacks different numbers of frames w.r.t. the model performance on each class.
% They focused on frame-level augmentation since not all frames are informative for class label.
% However, as the stacked feature at the snippet level does not provide a solution to overcome the weakly labeled problem and improve the quality of representations, their gains were limited.
However, as the stacked feature at the snippet-level does not directly address the weakly labeled problem and improve the quality of representations, their gains were limited.
Upon our findings to achieve gains in VLTR, we state that data imbalance in the video domain should be handled at the video-level simultaneously with balancing the majority-biased training.

%% file: 6_conclusion.tex
\section{Conclusion}
VLTR has the objective of learning from vastly accumulated real-world video streams that is difficult to balance between classes.
In this work, we presented three challenges in VLTR and ways to overcome each challenge.
For the limitation of task-irrelevant features and video-level supervision, we studied learnable feature aggregators to effectively extract video-level representation for the downstream task.
To minimize the information loss during compression, we utilized the local and global relationships by combining self- and codebook-attentive aggregators.
We then modeled the balanced behavior of the neural network towards discriminating between classes with minority-oriented vicinity expansion. 
To investigate each component, we carefully studied our components and showed the effects of each design.
Finally, by combining our components, we improved considerably over previous approaches for VLTR.

% Limitation
%\textbf{Limitation.}~While we focus on summarizing and addressing the challenges in VLTR, we have not studied how our approach works on the balanced distribution. Without the calibrated interpolation, we believe diversifying representations for all classes can be effective for video recognition tasks.
% \textbf{Limitation.}~While we focus on summarizing and addressing the challenges in VLTR, we have not extended our study to balanced distributions. Without differing the behaviors between classes, we believe diversifying representations for all classes can be effective for general video recognition tasks.

% \textbf{Potential negative societal impacts.}~With MOVE, users with malicious purposes can easily train the real world video dataset, that naturally tends to be vast and imbalanced, by simply adopting any pretrained networks. No extra foreseeable negative impacts are expected other than vicious purposes.

% \textbf{Potential negative societal impacts.}~As real-world video data naturally tends to follow long-tailed distributions, practitioners can easily train the video data by simply adopting the pretrained networks with malicious purposes. Other than that, there are no foreseeable negative impacts.
% 1. No boost with Temporal consistency
% 2. no extra dataset
% 3. extra -> intra 가 그냥 dataset에도 좋을수도있는데 test안해봄.

% Societal Impact 
% 1. real-world data tends to follow long-tailed class distribution, our work has 

%% file: appendix_0_overallflow.tex
\section{Overall Training Flow.}
The overall training flow is illustrated in Fig.~\ref{Fig.trainingflow}.
Here, we briefly summarize the procedure with algorithm~\ref{algorithm_MOVE}.
First, task-irrelevant snippet-wise features are dynamically sampled according to their class frequency and forwarded to aggregators.
Within aggregators, they are transformed to represent the video with class-discriminative information in self-attentive and codebook-attentive ways.
However, these features are not optimal to balance the biased training.
Hence, MOVE accordingly manipulates both the feature and the label space by sequentially densifying and enlarging the minority distribution with extrapolation and interpolation.
\begin{algorithm}[H]
    \caption{Training MOVE}
    \label{algorithm_MOVE}
    \begin{algorithmic} 
        \REQUIRE Dynamic Frame Sampler (DFS)
        \REQUIRE Smoothed distribution by Dyanmic Extrapolation $p_{ex}$
        \REQUIRE Balanced distribution by Calibrated Interpolation $p_{in}$
        \REQUIRE Classifier $f_{\theta}$, Self-attentive Aggregator $g_{\phi}$, Codebook-attentive Aggregator $h_{\psi}$
        \REQUIRE Input batch $\mathbf{X}$, Label batch $\mathbf{Y}$ 
        \REQUIRE Vector with number of samples for each class $\mathbf{n}$
        \REQUIRE Shuffled batch index set $\mathbf{s}$
        % \FOR{$i \in $\{1,..., \text{epoch}\}}
        \STATE $\begin{aligned}
	        & U, V = \text{DFS}(\mathbf{X}, \mathbf{n}) \\
	        & {\color{gray}{\triangleright \text{Phase} \textbf{1}: \text{Learnable Feature Aggregator}}} \\
		    & U_{sa}, V_{sa} = g_{\phi} (U, V) \;\{\text{Self-attentive Aggregator}\}\\
            & U_{ca}, V_{ca} = h_{\psi}(U, V) \;\{\text{Codebook-attentive Aggregator}\}\\
		    & U = U_{sa} \oplus U_{ca}, V = V_{sa} \oplus V_{ca} \\
            & {\color{gray}{\triangleright \text{Phase} \textbf{2}: \text{Minority-Oriented Vicinity Expansion}}} \\
            & \mathbf{Z} = p_{ex}(U, V, \mathbf{Y})  \;\{\text{Dynamic Extrapolation}\} \\
            & \mathbf{Z}, \mathbf{\hat{Y}} = p_{in}(\mathbf{Z}, \mathbf{Z}[\mathbf{s}], \mathbf{Y}, \mathbf{Y}[\mathbf{s}]) \;\{\text{Calibrated Interpolation}\} \\
		    & \hat{Z} = f_{\theta}(Z) \\
		    & \ell = \mathcal{L}_{BCE}(\mathbf{\hat{Z}}, \mathbf{\hat{Y}}) \\
		    & \theta \leftarrow \theta - \eta \nabla_{\theta} \ell \\
		    & \phi \leftarrow \phi - \eta \nabla_{\phi} \ell \\ 
		    & \psi \leftarrow \psi - \eta \nabla_{\psi} \ell
	    \end{aligned}$
    \end{algorithmic}
\end{algorithm}

%% file: appendix_1_implementation.tex
\section{Implementation details.}
% For all experiments regarding VideoLT and Imbalanced-MiniKinetics200 datasets, we follow the setups, e.g., model architecture and fixed seed number, from our baseline~\cite{videolt} for a fair comparison.
For a fair comparison, we follow the setups, e.g., model architecture and fixed seed number, from our baseline~\cite{videolt} for all experiments conducted in the paper.
To elaborate, we trained our models for 100 epochs with the binary cross-entropy function.
Adam optimizer is used with the learning rate initially set to 0.001 and decayed by the factor of 0.1 at the 50-th epoch.
For the hyperparameters in each aggregator, the number of heads and clusters are set to 4 and 64, respectively.
We use 2.0, 0.5, and 3 for each $\alpha$, $\gamma$, and $\sigma$.
For the trials, we use PyTorch as the machine learning framework and run on a single V100 NVIDIA GPU.

%% file: appendix_2_ablation.tex
\section{Tail-weighted criterion} 
In this section, we futher conduct ablation study on tail-weighted criterion.
Specifically, we investigate with varying strategies in Tab.~\ref{Tab.criterion}.
Except for the linear criterion that does not reflect dramatic gaps between number of samples in each category in the long-tailed distribution, all criteria regarding the number of samples are shown to boost performances.
Among various criteria, it is clear that the naive use of number of samples fits well for the criterion.
\begingroup
\setlength{\tabcolsep}{3.1pt} % Default value: 6pt
\renewcommand{\arraystretch}{1.} % Default value: 1
\begin{table}[h]
\centering
\scriptsize
\begin{tabular}{c|cccccc}
\hlineB{2.5}
% \multicolumn{3}{c|}{Method}& \multicolumn{6}{c}{ResNet-101} \\ \hline
Methods & All & H & M & T & A@1 & A@5 \\ \hlineB{2.5}
$\mathbf{n}_{s}$ & \textbf{0.719} & \textbf{0.815} & \textbf{0.753} & \textbf{0.644} & \textbf{0.730} & \textbf{0.883} \\ 
linear & 0.699 & 0.785 & 0.734 & 0.625 & 0.707 & 0.865 \\
$log({\mathbf{n}_{s}}) / \sum_{i=1}^{S}log({\mathbf{n}_i})$ & 0.712 & 0.805 & 0.747 & 0.637 & 0.723 & 0.878 \\
$\sqrt{\mathbf{n}_{s}} / \sum_{i=1}^{S}\sqrt{\mathbf{n}_i}$ & 0.717 & 0.811 & 0.751 & 0.643 & 0.728 & 0.880 \\
$log(\sqrt{\mathbf{n}_{s}}) / \sum_{i=1}^{S}log(\sqrt{\mathbf{n}_i})$ & 0.712 & 0.807 & 0.746 & 0.638 & 0.724 & 0.876 \\
\hlineB{2.5}
\end{tabular}
% \vspace{-0.1cm}
\caption{Different strategies to derive $\mathbf{q}$, that is used to create tail-weighted criterion in Eq.~\ref{eq_numsample}. $1_{st}$ column shows how the $s$-th element in $\mathbf{q}$ is assigned. $\mathbf{n}$ is the vector consisting of number of samples for each class. These vectors are normalized to be $\boldsymbol{\tau}$. 'linear' indicates a uniformly sampled vector between [0, 1] according to data frequency in each class.}
\label{Tab.criterion}
\end{table}
\endgroup
% we investigate how the performance changes as we choose varying strategies for the tail-weighted criterion $\tau$. 
% \textbf{Tail-weighted criterion.}
% In Tab.~\ref{Tab.criterion}, we investigate how the performance changes as we choose varying strategies for the tail-weighted criterion $\tau$. 
% Except the linear criterion that does not reflect dramatic gaps between number of samples in each category in the long-tailed distribution, all criteria regarding the number of samples are shown to boost the performances.
% Among various criteria, it is clear that the naive use of number of samples fits well for the criterion.

%% file: appendix_3_data.tex
\begin{figure*}
    \centering
    \includegraphics[width=1\textwidth]{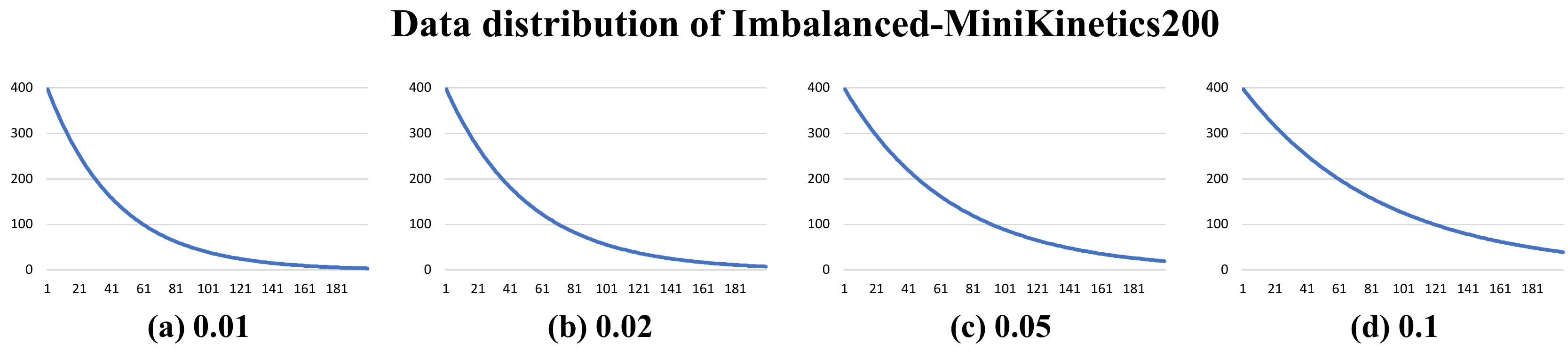}
    \caption{
    Data distribution according to varying imbalance ratios. (a), (b), (c), and (d) show the long-tailed distribution when the imbalance ratio is set to 0.01, 0.02, 0.05, and 0.1. The X-axis stands for video classes and Y-axis indicates the number of samples for each class.
    }
    \label{fig_minikinetics_distribution}
\end{figure*}

\section{Imbalanced-MiniKinetics200}
In contrast to the potential in its practicality of Video long-tailed recognition, we found that the benchmark is limited to only one large-scale dataset, VideoLT.
Particularly, the class frequency in the VideoLT dataset is shown to have linearity in logarithmic coordinate system~\cite{videolt} while other datasets collected from real applications may follow different data distribution.
Therefore, we found a need for a new benchmarking dataset in which the data distribution can be adjusted to simulate various long-tailed data circumstances.

In this regard, we synthetically created the Imbalanced-MiniKinetics200 dataset to facilitate the research for video long-tailed recognition.
To produce the Imbalanced-MiniKinetics200 dataset, videos belonging to each of 200 categories are sampled up to 400 videos from the Kinetics-400 dataset~\cite{kinetics-400} as similarly as \cite{xie2017rethinking}.
Note that due to the unavailability of some videos in Kinetics-400 and the fact that not all videos are required to create an imbalanced dataset, not all 80,000 videos are downloaded.
For details, we publicize the video list of downloaded videos in the code appendix.
Class names are in the next paragraph in the sorted order from the head to the tail.
Then, we used imbalance ratio $\mu$ to manipulate the harshness in imbalanced setting~\cite{cui2019class, cao2019learning} by generating a imbalance vector $v \in (\mu, 1)^K$ where K denotes the number of classes.
Specifically, $v$ is multiplied class-wisely as $n_i = n_{max}v_i$ where $n_i$ and $n_{max}$ indicate number of samples for i-th class and the maximum number of videos in certain class.
Depending on each scenario where imbalance ratio $\mu$ is set to different values in Tab.~\ref{Tab.Minikinetics}, we visualize how harsh long-tailed distribution is formed in Fig.~\ref{fig_minikinetics_distribution}.

In our github~\footnote{https://github.com/wjun0830/MOVE}, we also provide instructions about the procedure to create the Imbalanced-MiniKinetics200 dataset.
By following the instructions with specified video lists, the Imbalanced-MiniKinetics200 dataset can easily be downloaded.
Training codes to train and evaluate various imbalanced scenarios are also available, while different scenarios can be easily designed by modifying the imbalance ratio.

Below, we enumerate the classes used for the Imbalanced-MiniKinetics200 in order from head to tail classes.

\noindent\{
{0: 'beatboxing', 1: 'finger snapping', 2: 'air drumming', 3: 'country line dancing', 4: 'snatch weight lifting', 5: 'high jump', 6: 'cheerleading', 7: 'deadlifting', 8: 'spinning poi', 9: 'bowling', 10: 'lunge', 11: 'passing American football (not in game)', 12: 'dancing ballet', 13: 'dancing macarena', 14: 'gymnastics tumbling', 15: 'shot put', 16: 'kicking field goal', 17: 'giving or receiving award', 18: 'breakdancing', 19: 'arm wrestling', 20: 'headbanging', 21: 'shaking head', 22: 'golf driving', 23: 'sticking tongue out', 24: 'hurling (sport)', 25: 'golf putting', 26: 'catching or throwing baseball', 27: 'blowing glass', 28: 'brushing teeth', 29: 'catching or throwing softball', 30: 'front raises', 31: 'bookbinding', 32: 'eating spaghetti', 33: 'clean and jerk', 34: 'crawling baby', 35: 'juggling balls', 36: 'barbequing', 37: 'bench pressing', 38: 'flying kite', 39: 'bungee jumping', 40: 'feeding goats', 41: 'side kick', 42: 'contact juggling', 43: 'milking cow', 44: 'making snowman', 45: 'eating ice cream', 46: 'hitting baseball', 47: 'somersaulting', 48: 'capoeira', 49: 'dribbling basketball', 50: 'busking', 51: 'dunking basketball', 52: 'catching or throwing frisbee', 53: 'blowing out candles', 54: 'diving cliff', 55: 'hammer throw', 56: 'javelin throw', 57: 'high kick', 58: 'ice skating', 59: 'brushing hair', 60: 'cutting watermelon', 61: 'hula hooping', 62: 'dancing gangnam style', 63: 'archery', 64: 'abseiling', 65: 'baking cookies', 66: 'singing', 67: 'driving tractor', 68: 'curling hair', 69: 'eating burger', 70: 'long jump', 71: 'smoking hookah', 72: 'situp', 73: 'folding napkins', 74: 'cleaning floor', 75: 'shuffling cards', 76: 'jumping into pool', 77: 'biking through snow', 78: 'laughing', 79: 'feeding birds', 80: 'balloon blowing', 81: 'belly dancing', 82: 'ski jumping', 83: 'cooking chicken', 84: 'climbing tree', 85: 'chopping wood', 86: 'dying hair', 87: 'driving car', 88: 'feeding fish', 89: 'canoeing or kayaking', 90: 'sled dog racing', 91: 'shoveling snow', 92: 'doing nails', 93: 'petting animal (not cat)', 94: 'mowing lawn', 95: 'crying', 96: 'crossing river', 97: 'opening present', 98: 'smoking', 99: 'shaving head', 100: 'making pizza', 101: 'folding paper', 102: 'playing accordion', 103: 'braiding hair', 104: 'salsa dancing', 105: 'jetskiing', 106: 'shearing sheep', 107: 'slacklining', 108: 'filling eyebrows', 109: 'sharpening pencil', 110: 'playing badminton', 111: 'picking fruit', 112: 'passing American football (in game)', 113: 'throwing discus', 114: 'squat', 115: 'ice climbing', 116: 'skateboarding', 117: 'massaging back', 118: 'marching', 119: 'surfing crowd', 120: 'kitesurfing', 121: 'spray painting', 122: 'paragliding', 123: 'playing bagpipes', 124: 'parasailing', 125: 'tap dancing', 126: 'skiing (not slalom or crosscountry)', 127: 'pole vault', 128: 'playing basketball', 129: 'presenting weather forecast', 130: 'tai chi', 131: 'playing ukulele', 132: 'stretching leg', 133: 'tobogganing', 134: 'playing ice hockey', 135: 'waxing chest', 136: 'playing bass guitar', 137: 'playing cricket', 138: 'playing didgeridoo', 139: 'riding elephant', 140: 'motorcycling', 141: 'playing cello', 142: 'playing paintball', 143: 'waxing legs', 144: 'playing chess', 145: 'robot dancing', 146: 'playing poker', 147: 'snowkiting', 148: 'pull ups', 149: 'playing recorder', 150: 'playing xylophone', 151: 'playing tennis', 152: 'washing dishes', 153: 'riding or walking with horse', 154: 'playing volleyball', 155: 'swimming breast stroke', 156: 'playing clarinet', 157: 'roller skating', 158: 'reading book', 159: 'playing violin', 160: 'playing harmonica', 161: 'playing saxophone', 162: 'playing squash or racquetball', 163: 'playing harp', 164: 'tapping guitar', 165: 'rock climbing', 166: 'snowboarding', 167: 'playing trumpet', 168: 'throwing axe', 169: 'washing feet', 170: 'playing guitar', 171: 'scrambling eggs', 172: 'playing drums', 173: 'swimming backstroke', 174: 'riding unicycle', 175: 'punching bag', 176: 'walking the dog', 177: 'surfing water', 178: 'pushing car', 179: 'snorkeling', 180: 'trapezing', 181: 'tango dancing', 182: 'sailing', 183: 'pushing cart', 184: 'playing trombone', 185: 'weaving basket', 186: 'triple jump', 187: 'pumping fist', 188: 'washing hands', 189: 'scuba diving', 190: 'tying knot (not on a tie)', 191: 'welding', 192: 'water skiing', 193: 'trimming or shaving beard', 194: 'using computer', 195: 'zumba', 196: 'yoga', 197: 'wrapping present', 198: 'windsurfing', 199: 'unboxing'}
\}
%           \end{minipage}}
% \end{center*}